\documentclass{article}
\usepackage{float}

\PassOptionsToPackage{numbers}{natbib}
\usepackage[main, final]{neurips_2026}
\usepackage{placeins}
\usepackage[utf8]{inputenc} 
\usepackage[T1]{fontenc}    
\usepackage{hyperref}       
\usepackage{url}            
\usepackage{booktabs}       
\usepackage{amsfonts}       
\usepackage{nicefrac}       
\usepackage{microtype}      
\usepackage{xcolor}         
\usepackage{url}
\usepackage{hyperref} 
\usepackage{graphicx}
\usepackage{amsmath}

\usepackage{pdfpages,caption}

\usepackage{booktabs, tabularx, array}

\title{GramStatTexNet: Efficient, Interpretable, and Neuro-Inspired Texture Analysis-by-Synthesis}

\author{
  Vasha DuTell$^{1,2}$ \And Zeyu Yun$^{2}$ \And Anne Harrington$^{1}$ \And Michael Y.-S. Fang$^{4}$ \And Mark Hamilton$^{1}$ \And Christian K\"{o}vesdi$^{1}$ \And Edward H. Adelson$^{1}$ \And William T. Freeman$^{1}$ \And Ruth Rosenholtz$^{1,3}$ \\
  \\
$^1$MIT CSAIL \quad $^2$UC Berkeley \quad $^3$NVIDIA \quad $^4$Spectacular Labs \\
\texttt{\{vasha\}@berkeley.edu} \\
}

\begin{document}

\maketitle

\begin{abstract}


Deep networks achieve high realism in texture synthesis but often rely on opaque, over-parameterized representations that have limited direct connection to biological vision. Neuroscience-informed pyramid models are interpretable, compact, and biologically inspired, but rely on small, hand-curated statistics sets that are difficult to generalize or extend. We introduce GramStatTexNet, a hybrid analysis-by-synthesis framework that combines the multi-scale Gabor filter structure of classical models with the flexibility of Gramian-based correlations, organized into structured filter families. This structured representation enables analyses that are challenging for deep-feature methods: each statistic carries a named filter-family identity, allowing us to decompose synthesis fidelity into per-family contributions and to learn an interpretable, low-dimensional embedding that preserves perceptual texture content. We compare our framework to classical and deep texture-synthesis baselines across diverse texture categories, and on multiple perceptual metrics, highlighting the compactness and interpretability advantages of this hybrid representation, alongside competitive synthesis quality. This same framework extends naturally to peripheral textures via spatial pooling, and to dynamic textures using a spatiotemporal filter bank. GramStatTexNet provides a unified, interpretable framework for analyzing and modeling visual information with natural extensions across space, time, and eccentricity. Project page: \url{https://vdutell.github.io/gstn}

\end{abstract}

\section{Introduction}

The human visual system demonstrates impressive performance and efficiency at processing and interpreting the dynamic visual world. In particular, human vision has the ability to recognize complex textures that are essential for identifying objects, surfaces, and materials. Humans achieve this both for static scenes and in the context of motion, enabling us to navigate and interact seamlessly with our environment. Understanding how to model these capabilities is crucial for advancing applications in computer vision and machine learning. These goals are also closely related to interpretability, where biological-vision can serve as a natural source of inspiration for learning effective and more transparent visual representations.  



Neuroscience-informed approaches to texture synthesis (and closely related peripheral vision) often model visual areas using non-parametric multi-scale pyramids, decomposing images into filter responses, and aggregating features into statistics sets \citep{portilla2000parametric}. These statistics sets are predictive of neural responses to natural scenes \citep{ziemba2016selectivity}, produce perceptually plausible textures, and serve as models of peripheral vision when applied over spatial pooling regions \citep{rosenholtz2012summary,freeman2011metamers}. Despite the success of such models for texture images, they rely on hand-curated statistics sets, which are challenging to design and to extend. One example of this is the spatiotemporal domain, where spatial statistics often lack clear analogues in time. 

Deep-learning based approaches to texture and peripheral synthesis offer strong results in terms of quality and speed, and eliminate the need for hand-curated features, but their reliance on over-parameterized latent spaces comes at the cost of interpretability and biological feasibility. Gram-matrix-based approaches synthesize textures by matching correlations between filter responses, using either deep multi-layer networks \citep{gatys2015texture} or single-layer random filters \citep{ustyuzhaninov2017does}. This approach has also been successful for generating peripheral syntheses \citep{wallis2017towards,deza2017towards}. However, these models rely on pretrained layers from networks like VGG \citep{simonyan2014very}, and thus offer no representational compression compared to the highly efficient texture models from visual neuroscience.


In this work, we introduce GramStatTexNet (GSTN), an analysis-by-synthesis approach that bridges interpretability, biological plausibility, and synthesis quality in texture modeling. Our model generates texture syntheses comparable in quality to deep network-based models while remaining interpretable and efficient in representation size. We organize statistical representations into structured families and use contrastive learning to identify the most critical statistics for texture differentiation, reducing representation size while minimizing synthesis quality reduction and exposing which spatial-frequency, orientation, and color correlations carry texture identity, yielding testable hypotheses about which correlations visual cortex may encode. We further demonstrate that the same framework is compatible with classical peripheral-vision synthesis pipelines via spatial pooling, and extends naturally to dynamic video textures. Together, GramStatTexNet provides a unified, interpretable framework for modeling visual information that operates across space, time, and eccentricity.

\section{Previous Work}

Early approaches to texture synthesis were deeply rooted in efforts to model human vision both in neuroscience and computer vision. Julesz' \citep{julesz1962visual} early work utilized N-th order statistics in pixel space. Texture representation in Fourier space \citep{matsuyama1983structural}, and later in pyramid space \citep{burt1987laplacian,freeman1991design} improved texture models significantly \citep{heeger1995pyramid}. Along with considerable improvement in synthesis quality, came a strong link to the neuroscience of human visual processing, with the oriented multi-scale filters of the steerable pyramid \citep{simoncelli1995steerable} mirroring both the receptive fields of neurons found in area V1 \citep{turner1986texture, malik1990preattentive}, as well as their response properties \citep{olshausen1997sparse}. The Portilla \& Simoncelli (P\&S) model \citep{portilla2000parametric} further improved synthesis quality by including pixel, autocorrelation, and magnitude statistics on pyramid responses.

For video, early methods resampled frames of the input \citep{Schödl2000texture}, reproducing a clip without generating new spatiotemporal structure, or modeled temporal evolution with linear dynamical systems over a low-dimensional appearance model \citep{doretto2003dynamic}. In vision science, motion has long been modeled with spatiotemporally oriented filters \citep{adelson1985spatiotemporal, simoncelli1993distributed}, and oriented spacetime energies have been used to represent and recognize dynamic textures \citep{derpanis2010dynamic, derpanis2012spacetime}. Yet the compact pyramid-plus-correlation tradition of \citet{portilla2000parametric} has not, to our knowledge, been extended to temporal synthesis, despite extensive characterization of human spatiotemporal sensitivity \citep{kelly1979motion, krajancich2021perceptual}.

Summary statistics texture models have been extended as peripheral vision models, with the incorporation of spatial pooling \citep{rosenholtz2012summary, freeman2011metamers}. This method involves synthesizing texture locally in many small overlapping regions, which when optimized jointly, lead to full-field image syntheses. By progressively increasing the size of these regions with increased distance from central vision, these models reflect loss of information in peripheral vision beyond the simple blurring of photoreceptor density loss. As analysis-by-synthesis models, they enable the synthesis of `metamers' or `mongrels'. These image syntheses appear lossy when viewed foveally (with central vision), but appear normal when viewed peripherally (metamers), and have been shown to capture the information available to human peripheral vision, therefore predicting human performance on a wide range of visual tasks (mongrels). These models have been since extended to incorporate color \citep{wallis2017towards}, GPU acceleration \citep{brown2021efficient}, expanded field of view \citep{broderick2023foveated}, and have been combined with deep learning to explore higher-level similarities and divergences in human and machine vision \citep{harringtoncoco}.

In addition to neural-inspired approaches, pure computer vision and deep learning approaches have had success in realistic synthesis of both static and dynamic textures. In the spatial domain, state-of-the-art texture synthesis often uses diffusion models to generate photorealistic results \citep{chen2023text2tex, youwang2024paint}, with the advantage of controllability through text prompting. In addition, neural cellular automata enable ultra-compact spatial \citep{mordvintsev2021mu} and real-time controllable spatiotemporal \citep{pajouheshgar2023dynca} synthesis. Deep features have also been used to evaluate texture properties directly, such as tileability \citep{rodriguez2024textile}. For dynamic textures, long-range spatial and temporal dependencies have been shown to improve syntheses \citep{zhang2021dynamic}.

Closest to our approach, a line of work in deep learning generalizes these classical summary statistics to learned features \citep{gatys2015texture, tesfaldet2018two}. Style transfer \citep{gatys2016image} is the extension of a spatial texture method \citep{gatys2015texture} which synthesizes textures by matching summary statistics defined as correlations between layers of pre-trained image networks such as VGG-Net \citep{simonyan2014very}, with even a single layer of random filters sufficing \citep{ustyuzhaninov2017does}. Like the classical models, this approach has been extended to peripheral vision \citep{deza2017towards, wallis2017towards} and, with two-stream architectures, to video \citep{tesfaldet2018two}. On one hand, these methods demonstrate strong results in terms of synthesis quality while eliminating the need for hand-curated statistics sets. However, compared to pyramid-based models they have three major disadvantages: 1) They are highly-overparametrized, with orders of magnitude more statistics. 2) They replace human visual neuroscience-informed filters with less biologically-feasible convolutional layers. 3) Their summary statistics of deep network hidden layer correlations are difficult to interpret.

\section{GramStatTexNet}

\begin{figure*}[ht]
  \centering
   \includegraphics[width=0.99\textwidth]{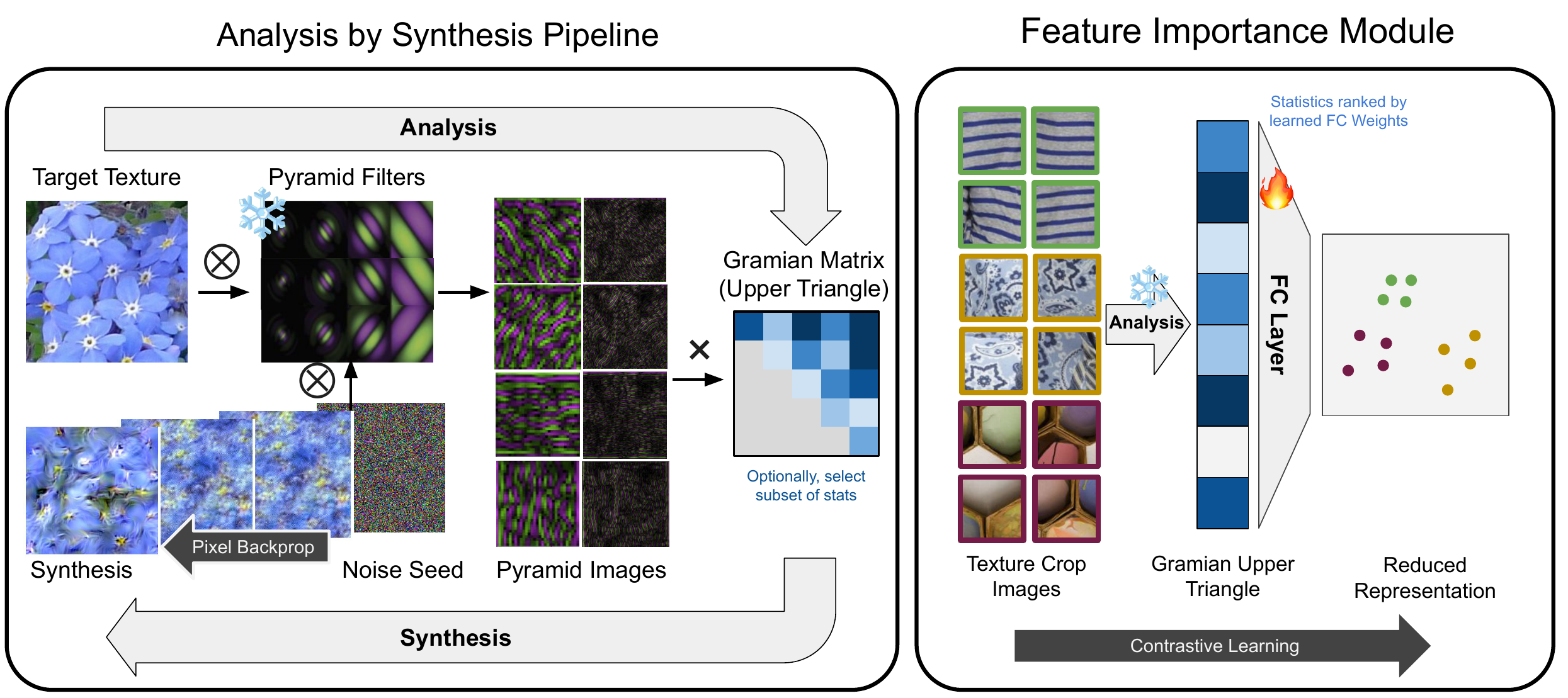}
  \caption{\textbf{Left:} GramStatTexNet, our analysis-by-synthesis texture model, decomposes images into stacks (pyramids) of varying scales, orientations, and color channels, using a multi-scale pyramid filter bank. Pair-wise correlations between all pyramid images are represented in the Gram matrix; the upper triangle of this Gram matrix represents the full statistics set. Synthesis proceeds by calculating loss on statistics between a target and noise seed image, backpropagating this loss through the network to adjust pixel values until a synthesized image is produced.  \textbf{Right:} To learn a reduced statistics set, a single fully connected layer compresses the Gram upper triangle to a reduced representation. These weights are trained using a contrastive loss, self-supervised by the objective of crops from the same texture image lying nearby in the reduced feature space.}
  \label{fig:model}
\end{figure*}

We present a simple but powerful model (Figure \ref{fig:model}) that combines the interpretable and biologically-inspired pyramid filters of \citep{portilla2000parametric} with the Gramian matrix representation of \citep{gatys2015texture}. GramStatTexNet (GSTN) leverages the GPU-accelerated multi-scale pyramids from \citep{brown2021efficient}, convolving input images with pyramid filters at individual combinations of orientation, phase, scale, and color, the output of which we call `pyramid images'. We treat each pyramid image as a channel, and calculate the pair-wise correlations between each pyramid image pair, collapsing over space as a Gram matrix. We use the upper triangle and diagonal of this matrix as the full statistics set for downstream analysis, where each statistic corresponds to a pairwise correlation between filter responses. This can be used not only to analyze the properties of a given texture, but also to synthesize new texture images. 

\FloatBarrier
\subsection{Texture Synthesis}

\begin{figure}[ht]
  \centering
  \includegraphics[width=\textwidth]{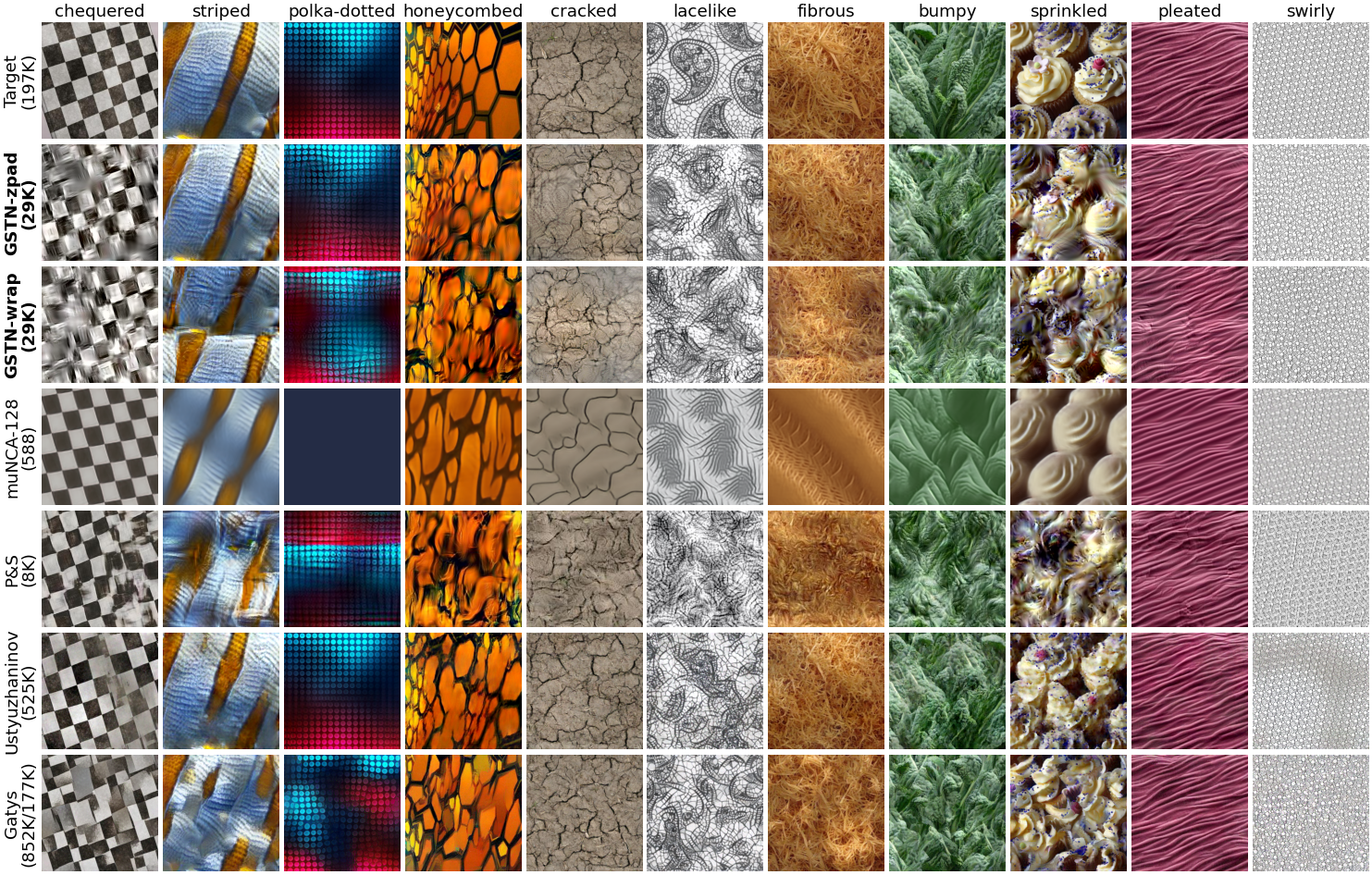}
  \caption{Texture syntheses across DTD \cite{cimpoi2014describing} categories spanning a range of structural regularity. Top row: target textures. GSTN matches the Gramian of multi-scale Gabor filters; GSTN-wrap uses circular (wrapped) pyramid convolutions, assuming a stationary, tileable texture and is faster, while GSTN-zpad zero-pads before the Fourier transform, so opposite edges do not interact and no tiling is assumed (Appendix~\ref{sec:wrapzpad}). $\mu$NCA~\citep{mordvintsev2021mu} is a small-footprint cellular automaton model. P\&S ~\citep{portilla2000parametric} uses a similar multi-scale Gabor strategy, but matches 8,400 hand-picked statistics rather than a full Gramian. Gatys~\citep{gatys2015texture} matches the Gramian of deep network VGG-19. Ustyuzhaninov ~\citep{ustyuzhaninov2017does}, like Gatys, matches a Gramian but calculated on random convolutional filters.}
  \label{fig:imsynths}
\end{figure}

\begin{figure}[ht]
  \centering
  \includegraphics[width=\textwidth]{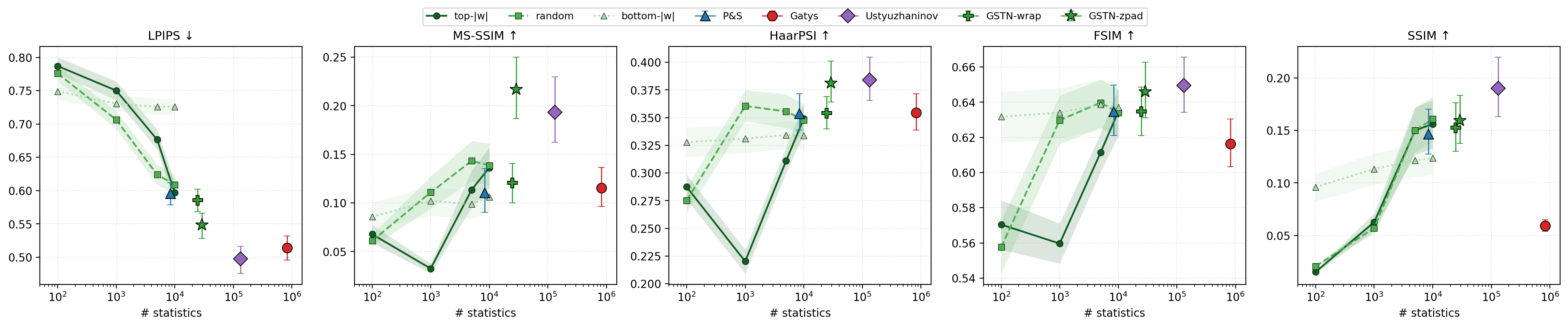}
  \caption{Pareto comparison: representation size versus synthesis quality across five perceptual metrics on 141 DTD validation textures. GSTN-zpad sits on the Pareto frontier between P\&S~\citep{portilla2000parametric} and the order-of-magnitude-larger feature-space methods including Ustyuzhaninov~\citep{ustyuzhaninov2017does} and Gatys~\citep{gatys2015texture}, achieving competitive quality with 30$\times$ fewer statistics than Gatys on four of five metrics. On LPIPS, Gatys is superior, consistent with the metric evaluating similarity in the same VGG features that Gatys' synthesis loss optimizes. Depletion variants (top-|w|, random, bottom-|w|) extend the GSTN frontier toward smaller representation sizes. $\mu$NCA not included due to compute limitations. See Appendix Figures \ref{fig:gstn_v_ps} and \ref{fig:gstn_v_gatys}. Error bars: 95\% bootstrap confidence intervals on the mean over 141 textures, 1{,}000 resamples.}
  \label{fig:pareto}
\end{figure}

We first apply our model in an analysis-by-synthesis framework to generate novel texture images, matching the statistics set of an input image, and synthesizing new textures by optimizing noise images to match the statistics set. We demonstrate good synthesis quality over a variety of texture types and regularities, sampling textures from a variety of categories from the Describable Texture Dataset (DTD) \citep{cimpoi2014describing} (Figure \ref{fig:imsynths}). Comparing GSTN to other related texture synthesis models, we see quality improvement over the P\&S \citep{portilla2000parametric} multi-scale-pyramid synthesis method, and competitive quality to the much larger, deep-network and convolution-based synthesis methods \citep{gatys2015texture, ustyuzhaninov2017does}, despite over an order of magnitude fewer statistics. Evaluating these results quantitatively, and including MS-SSIM~\citep{wang2003multiscale} (Figure \ref{fig:pareto}), we find that GSTN outperforms P\&S on the majority of categories across all nine evaluation metrics tested (30--46 of 47 categories per metric; Appendix Figure \ref{fig:gstn_v_ps}). Gatys ~\citep{gatys2015texture} is superior on deep-feature metrics (DISTS, LPIPS, DreamSim ~\citep{ding2020image,zhang2018unreasonable,fu2023dreamsim}), which share VGG features or are trained on object-centric similarity judgments, while GSTN is better performing on classical perceptual and structural metrics (FSIM, SSIM, VIF, GMSD, HaarPSI ~\citep{zhang2011fsim,wang2004image,sheikh2006image,xue2013gradient,reisenhofer2018haar}, PSD-L2) in 32--46 of 47 categories  (Appendix Figure~\ref{fig:gstn_v_gatys}). Overall, this indicates quality parity of GSTN with Gatys at a 30$\times$ smaller representation. In addition, Ustyuzhaninov's random-filter model~\citep{ustyuzhaninov2017does} has slightly higher quality but uses 18$\times$ more statistics, while $\mu$NCA~\citep{mordvintsev2021mu} is far more compact but must be trained per texture and has severe loss of detail.

\section{Interpretable Representation of Texture}

\subsection{Correlation Families}

\begin{table*}[!b]
\centering
\scriptsize
\setlength{\tabcolsep}{4pt}
\renewcommand{\arraystretch}{0.9}
\caption{Summary of statistical families in GSTN. Non-sub-band statistics contain at least one marginal, highpass, or lowpass statistic. Each sub-band statistic correlates two channels of the pyramid representation, with channel specified by phase, scale (level), orientation, and color. A family is defined by which of these attributes are shared and which differ between the two correlated channels. Families marked † are structured: only a single attribute varies, approximating a subset of the P\&S statistics. Unmarked families are unstructured, with multiple attributes varying at once. S = scale, C = color, O = orientation, P = phase (re = real, im = imaginary, mag = magnitude).}
\label{tab:families}
\begin{tabularx}{\textwidth}{@{} l l l l >{\raggedright\arraybackslash}X r @{}}
\toprule
Name & Code & Shared & Varies & Correlations & \# \\
\midrule
\multicolumn{6}{@{}l}{\textit{Non-sub-band}} \\
Marginal$^\dagger$ & \texttt{marginal} & -- & -- & Per-color pixel mean, variance, std (appended) & 9 \\
Highpass$^\dagger$ & \texttt{highpass} & -- & -- & Highpass residuals across three color channels & 6 \\
Lowpass$^\dagger$ & \texttt{lowpass} & -- & -- & Two lowpass channels differing in S, C, both, or neither & 231 \\
Non-oriented $\times$ band & \texttt{pass\_multi} & -- & -- & Highpass or lowpass statistic with any sub-band channel & 5,247 \\
\midrule
\multicolumn{6}{@{}l}{\textit{Sub-band}} \\
Cross-orientation$^\dagger$ & \texttt{sub\_Xori} & S, C, P & O & Differ only in orientation & 324 \\
Cross-color$^\dagger$ & \texttt{sub\_Xcolor} & S, O, P & C & Differ only in color & 432 \\
Cross-scale$^\dagger$ & \texttt{sub\_Xlevel} & C, O, P & S & Differ only in scale & 540 \\
Cross-phase, re $\times$ im$^\dagger$ & \texttt{sub\_Xri\_eq} & S, C, O & P (re, im) & Real and imaginary parts of the same filter output & 72 \\
Cross-phase, re $\times$ mag$^\dagger$ & \texttt{sub\_Xrm\_eq} & S, C, O & P (re, mag) & Real part and magnitude of the same filter output & 72 \\
Cross-phase, im $\times$ mag$^\dagger$ & \texttt{sub\_Xim\_eq} & S, C, O & P (im, mag) & Imaginary part and magnitude of the same filter output & 72 \\
Real, multi-attr. & \texttt{sub\_real\_Xmulti} & P (re) & S, C, O & Two real outputs differing in $\geq$2 of S, C, O & 2,196 \\
Imaginary, multi-attr. & \texttt{sub\_imag\_Xmulti} & P (im) & S, C, O & Two imaginary outputs differing in $\geq$2 of S, C, O & 2,196 \\
Magnitude, multi-attr. & \texttt{sub\_magn\_Xmulti} & P (mag) & S, C, O & Two magnitude outputs differing in $\geq$2 of S, C, O & 2,196 \\
Mixed-phase, multi-attr. & \texttt{sub\_Xmulti} & -- & P, S, C, O & Different phase types, differing in $\geq$1 of S, C, O & 15,336 \\
\midrule
\textbf{Total} & & & & & \textbf{28,929} \\
\bottomrule
\end{tabularx}
\end{table*}

An advantage of our method is that the Gramian representation for a given input texture is comprised of correlations between known pyramid filters, and is thus transparent and interpretable such that the properties of any given texture can be analyzed and understood from first-principles. In addition, through ablation, the quality of syntheses can be attributed to different statistical families. To make sense of these Gramian correlation statistics, we organize them into families falling into two main groups. First, statistics appending or correlating one or more non-sub-band (non-pyramid level) filter images, which we call `Non-sub-band', and includes marginal statistics. The second group we call `Sub-band' and computes correlations exclusively between sub-band images (Table \ref{tab:families}). We further organize the Sub-band group into sub-groups, based on which properties are shared and differ between the pyramid images correlated. The hand-curated statistics sets from  \citep{portilla2000parametric} rely heavily on correlation statistics analogous to the first 3 non-sub-band and first 6 sub-band families, which we call `structured' statistics. These share all properties except for one (e.g. Cross-scale at neighboring levels). Family codes mark the varying attribute with X (e.g., \texttt{sub\_Xori}: sub-band pairs differing only in orientation; \texttt{sub\_Xmulti}: pairs differing in several attributes), and each channel is indexed as \texttt{<phase>\_L<scale>\_O<orientation>\_C<color>}.

\begin{figure}[ht]
  \centering
  \includegraphics[width=\textwidth]{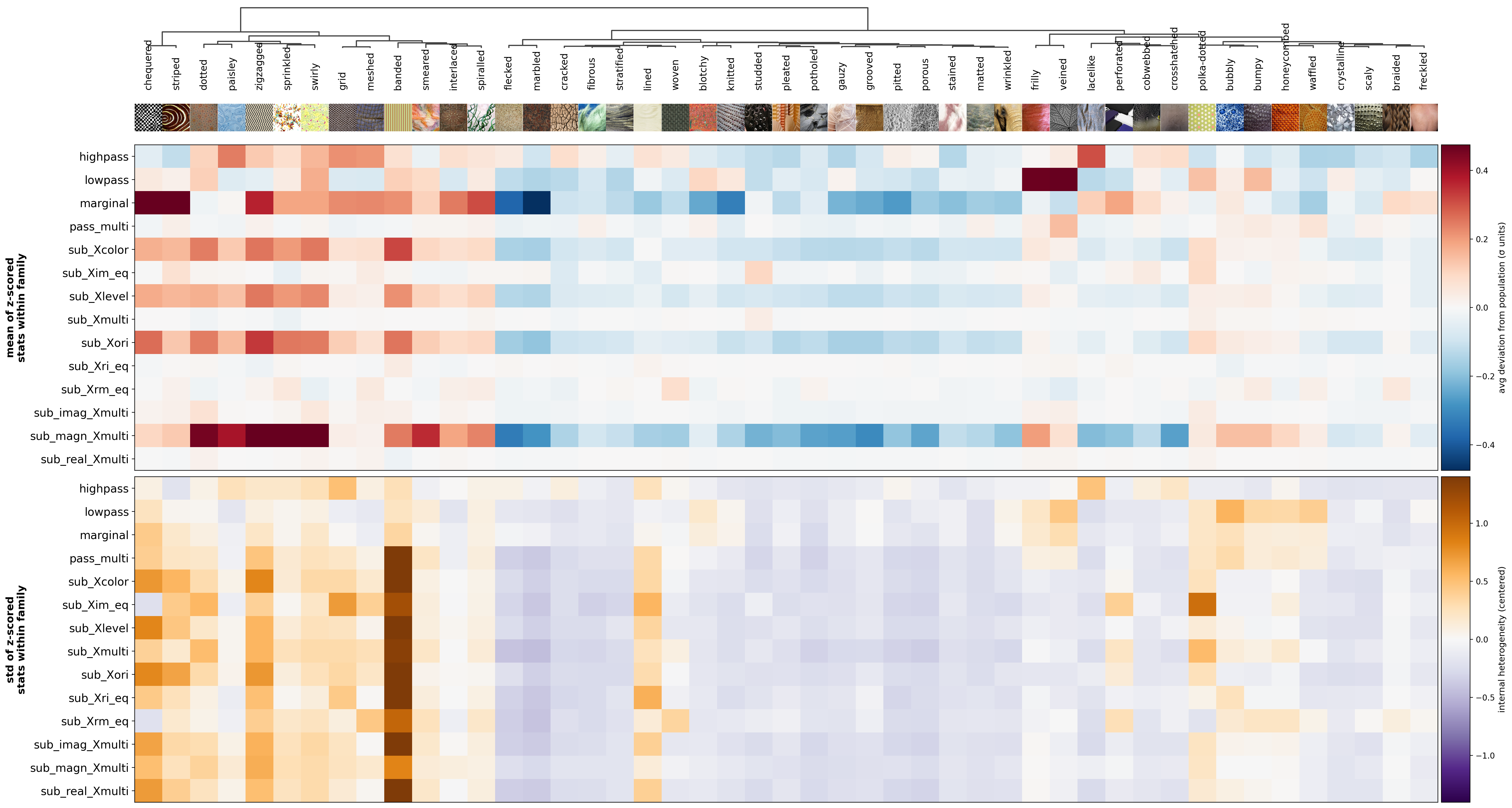}
  \caption{Family-level texture fingerprints across DTD. Top: Hierarchical clustering on per-texture category mean signatures recovers a smoothness vs structure ordering without supervision; one representative exemplar per category at each leaf. Top heatmap: per-category mean of within-family z-scored stat values. Bottom heatmap: per-category within-family std of z-scored stats (centered; orange =
  above-average within-family heterogeneity, purple = below).}
  \label{fig:texture_dendrogram}
\end{figure}

To analyze how statistical families differ across texture types, we group images by category \citep{cimpoi2014describing} and evaluate the per-category mean and within-category standard deviation of each family's z-scored correlations (Figure~\ref{fig:texture_dendrogram}). Unsupervised hierarchical clustering on the mean fingerprints recovers an interpretable global organization, with regular/structured textures (chequered, striped, polka-dotted, grid) on the left, balanced soft/irregular textures (frilly, veined, lacelike, bubbly) on the right, and stochastic mid-frequency textures (flecked, marbled, cracked, fibrous) in the middle.

Several categories show distinctive single-family signatures that map onto recognizable image properties: chequered has the strongest marginal deviation, likely reflecting that its pixels are mostly either fully dark or fully light; lined and banded show elevated Magnitude multi-attr. (sub\_magn\_Xmulti), capturing the repetition of a single dominant orientation across spatial scales; frilly and veined show strong lowpass elevation, reflecting their smooth, gradual variation rather than sharp edges; and crosshatched shows elevated highpass, reflecting the fine-scale edges produced by its overlapping line patterns. Three families (Cross-phase re $\times$ im (sub\_Xri\_eq), Cross-phase re $\times$ mag (sub\_Xrm\_eq), and Imaginary multi-attr. (sub\_imag\_Xmulti)) are nearly flat, indicating that they may carry little information; we test their causal relevance via family ablation.

The heterogeneity heatmap (Figure \ref{fig:texture_dendrogram}, bottom) shows that texture types banded, lined, and zigzagged are diverse in statistics across exemplars (corresponding to a broad range of stripe widths, orientations, and contrasts), while soft/irregular categories on the right (frilly, cobwebbed, freckled) are statistically tight despite visual diversity, suggesting the family-level summary captures a stable statistical regime even when individual exemplars of a texture type look quite different.

\begin{figure}[t]
  \centering
  \includegraphics[width=0.99\textwidth]{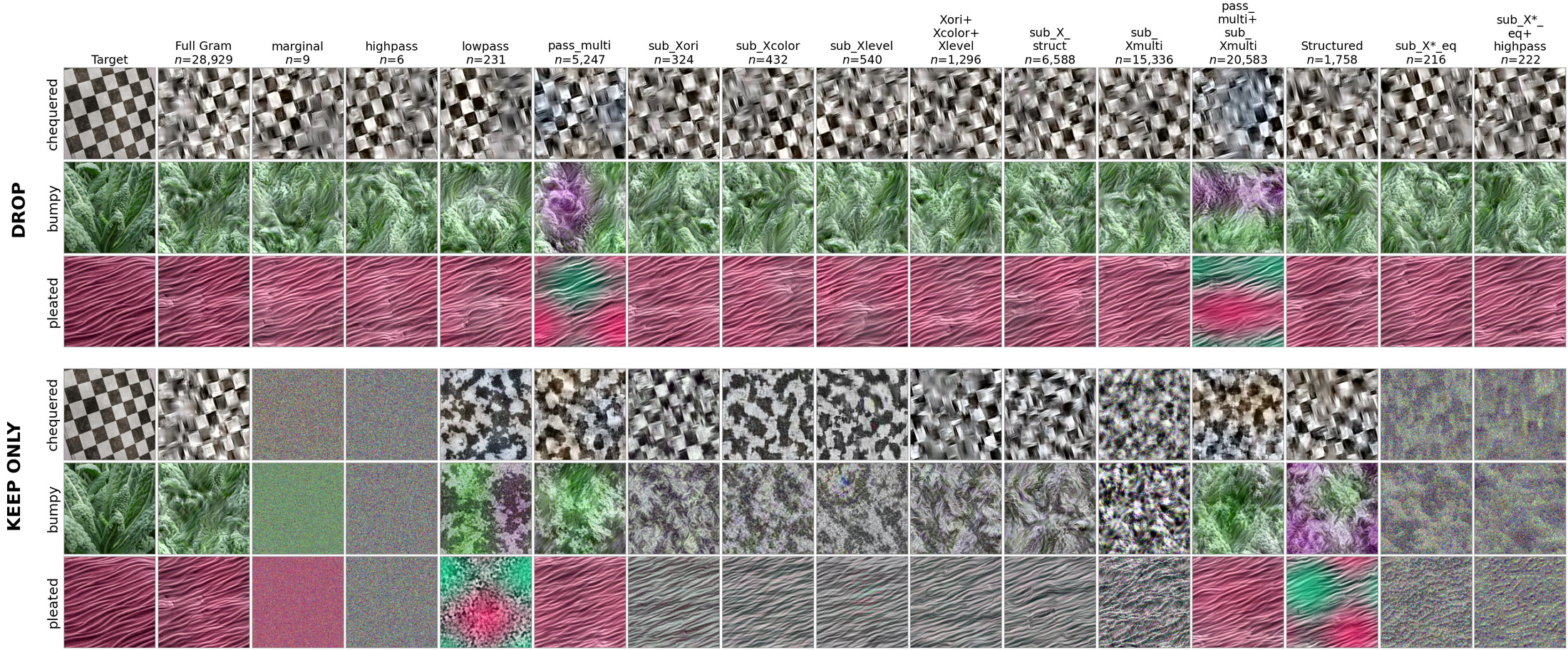}
  \caption{Ablated syntheses optimized using statistical subsets defined by categorical families; column codes follow Table~\ref{tab:families} ($+$ = union). sub\_X\_struct is the three single-phase multi-attribute families (Real, Imaginary, Magnitude), which are unstructured; sub\_X*\_eq is the three cross-phase families.}
  \label{fig:ablationfamilies}
\end{figure}

\subsection{Family-Based Ablation}

To understand how synthesis quality is related to different family groups, we perform paired ablation experiments, synthesizing textures with each family removed (DROP) and with each family alone (KEEP ONLY) (Figure~\ref{fig:ablationfamilies}). DROP demonstrates which families are necessary, and KEEP ONLY shows which are sufficient. We find that no single family is sufficient on its own. Even the largest individual family (Mixed-phase multi-attr., n=15,336) produces visibly degraded textures, and the union of \emph{structured} families related to classical pyramid statistics (n=1,758) preserves only coarse texture identity. The bulk of synthesis-relevant information is carried by the large \emph{unstructured} families whose correlations span multiple filter attributes simultaneously (Non-oriented $\times$ band (pass\_multi), Mixed-phase multi-attr. (sub\_Xmulti)). Interestingly, synths from the families with near-zero  values in Figure~\ref{fig:texture_dendrogram} (Cross-phase re $\times$ im (sub\_Xri\_eq), Cross-phase re $\times$ mag (sub\_Xrm\_eq), Imaginary multi-attr.(sub\_imag\_Xmulti)) show some of the worst quality, consistent with these  carrying little information. Conversely, dropping individual families produces only minor visual degradation in most cases, with even the Mixed-phase multi-attr. drop of 15,336 statistics causing little visible loss, and reflecting substantial redundancy in the statistical representation. One notable exception is that dropping Non-oriented $\times$ band visibly disrupts synthesis. Together, these results suggest that the classical structured-statistics approach captures only a fraction of synthesis-relevant content; the bulk of texture identity is encoded in cross-attribute correlations (Non-oriented $\times$ band, Mixed-phase multi-attr.) that are absent from classical pyramid models, and may explain their characteristic synthesis limitations. The structured families approximate a subset of the P\&S statistics, whose correlations drive V2 responses in macaque and human~\citep{freeman2013functional} and predict V4 selectivity and human texture discrimination ~\citep{okazawa2015image}. This grounding is at the level of families, not GSTN itself; the cross-attribute families are untested, making their importance a testable hypothesis about what higher visual areas encode beyond the classical set.

\section{Feature Importance Ordering}

To further investigate the contribution of individual texture statistics and families to synthesis quality, we use contrastive learning to reduce the correlation statistics to a compressed feature vector that can group textures (Figure \ref{fig:model}, right). We train a single fully connected layer (28,929x100) on the output of GSTN using InfoNCE \citep{oord2018representation} to group texture crops from the same parent texture in the compressed space. We find that the trained single layer is sufficient to group same texture crops in embedding space (Appendix Figure \ref{fig:umap}). We solve the contrastive learning problem using gradient-based optimization. However, in this case with a single-layer encoder, it can also be solved analytically; we derive this solution in Appendix \ref{sec:cl_derivation}.

\begin{figure}[ht]
  \centering
  \includegraphics[width=\textwidth]{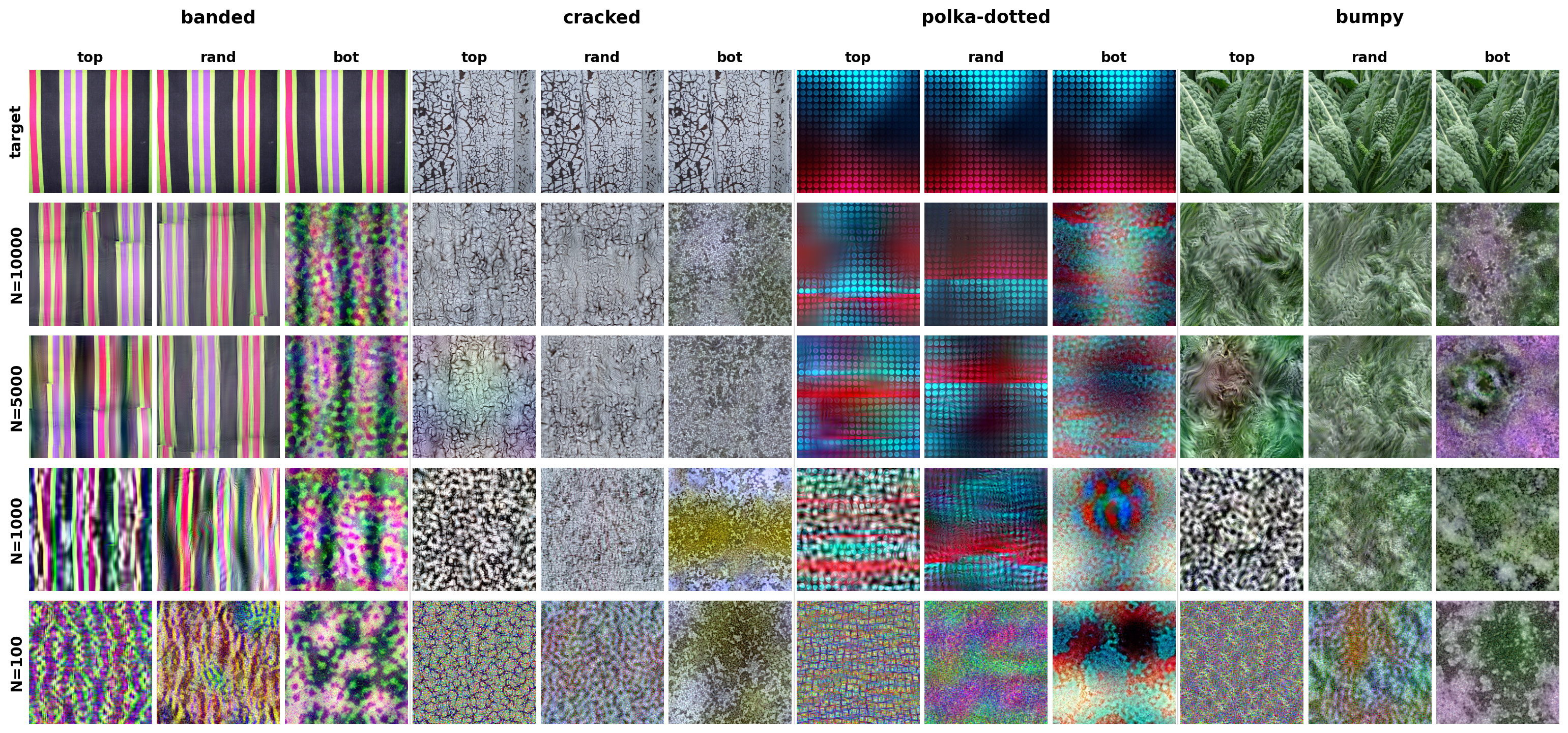}
  \caption{Stat-depletion comparison. Syntheses produced when only N stats are matched of the full 28,929-dimensional GSTN representation. Columns within each category compare three selection rules: top-N by trained-compressor importance, rand-N randomly sampled, and bot-N least-important. Synths use GSTN-wrap for computational efficiency.}
  \label{fig:ablationsynths}
\end{figure}

To evaluate the importance of different statistics to the contrastive learning task, we order statistics from most to least important based on both the absolute value of their weightings (Appendix Figure \ref{fig:weights}), as well as by their Shapley values \citep{roth1988shapley} (not shown, similar results). We find correlations between sub-band filter outputs with few or no shared attributes (Mixed-phase multi-attr.) are most important to the contrastive learning task. By contrast, correlations between lowpass filter outputs are least important for grouping textures. We visualize the relative contributions of individual families to the total statistics set when ordered by importance (Appendix Figure \ref{fig:familycontributions}), and find that Mixed-phase multi-attr., highpass, Magnitude multi-attr., and Non-oriented $\times$ band families make an out-sized contribution to the most important statistics.

To evaluate the transfer of importance in the contrastive task to synthesis quality, we perform a set of additional synthesis ablation experiments, matching only statistics of the highest weighting in the contrastive learning task (Figure \ref{fig:ablationsynths}). We find that importance in contrastive learning broadly transfers to synthesis quality: by selecting the top-ranked statistics, we achieve acceptable syntheses with only 5,000 of the total 28,929 statistics, an 83\% reduction. Statistics from bottom of the ranking, by contrast, produce extremely poor syntheses, confirming that the ranking carries real information about which statistics are needed for texture reproduction.

When we compare top-ranked statistics to randomly-selected control sets of the same size, we find that randomly chosen statistics in fact out-perform the top-ranked ones for synthesis (Figure \ref{fig:pareto}), even achieving acceptable synthesis with only 1000 statistics in some cases (Figure \ref{fig:ablationsynths}). We believe this reflects a property of the contrastive task, which selects statistics that discriminate between texture categories, and for which a few specific families (Mixed-phase multi-attr., highpass) maximally separate categories in the compressed embedding space. Synthesis, on the other hand, requires statistics that broadly constrain perceptual appearance across many aspects of a texture. A random subset of statistics samples broadly across the full feature space, including individually low-discriminability statistics that nonetheless contribute to constraining the synthesis to produce a coherent texture. The top-N subset may be more compact and more interpretable, but provides a narrower basis for reconstruction. This finding aligns with the classical distinction between discriminative and generative objectives in machine learning \citep{ng2001discriminative}: representations optimized to discriminate between categories are not necessarily optimal for modeling the joint statistical structure within them. This result also echoes \citep{ustyuzhaninov2017does}, where correlations on random convolutional filters prove sufficient for texture synthesis.




\section{Beyond Static Foveal Synthesis}

Having established an interpretable statistical representation for static textures, we now demonstrate that the same framework extends naturally beyond the static and uniform texture image setting. We present two extensions: applying the model over spatial pooling regions to produce foveated syntheses compatible with peripheral-vision pipelines, and replacing the spatial filter bank with a spatiotemporal one to synthesize dynamic textures. Neither extension requires changes to the underlying analysis-by-synthesis pipeline, only to the spatial aggregation of the model, or a temporal dimension to the filter bank; these illustrate the generality and interoperability of our pyramid plus correlation-statistics approach.

\begin{figure}[t]
  \centering
  \includegraphics[width=0.95\textwidth]{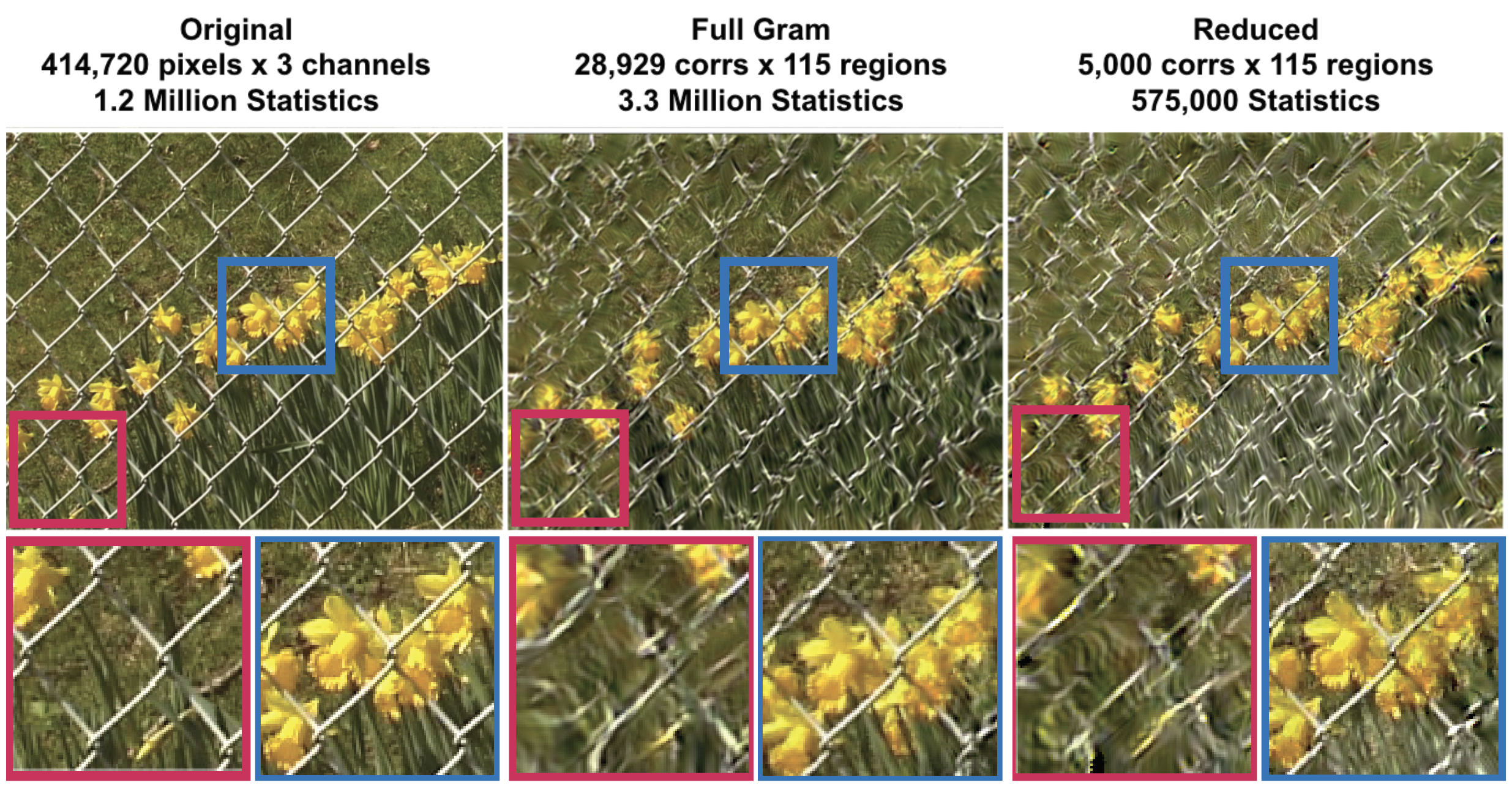}
  \caption{Spatial pooling produces foveated syntheses analogous to those used in peripheral-vision studies, where the center of the visual field is rendered at high fidelity, and the edges of vision are lossy. Original image (left) represents 1.2 Million values per image. Syntheses using the full model (middle) with 28,929 statistics per pooling region, and 115 pooling regions increases the representation size to 3.3 million, and creates quality syntheses. Our reduced model (right) decreases the statistical representation to 5,000 statistics per pooling region, for 575,000 total values, while still achieving good results.}
  \label{fig:pooled_ablations}
\end{figure}

\textbf{Peripheral Synthesis:} Demonstrating compatibility with peripheral-vision synthesis, we combine our method with spatial pooling following \citep{freeman2011metamers, rosenholtz2012summary} and motivated by evidence that pooled summary statistics predict crowding~\citep{balas2009summary}, generating foveated synthesis on full-size images which have non-uniform texture (Figure \ref{fig:pooled_ablations}). The resulting syntheses preserve high fidelity at gaze center and progressively introduce spatial distortions toward the periphery, matching the smooth, spatially-varying degradation pattern produced by prior peripheral synthesis pipelines \citep{freeman2011metamers, brown2021efficient}. Utilizing the reduced representation learned via contrastive learning, we reduce the size of our peripheral syntheses significantly, while retaining high quality. Our resolution study (Appendix Table~\ref{tab:resolution}, Figure~\ref{fig:resolution_grid}) quantifies the tradeoff for a fixed statistic budget with quality falling as region size grows, though regions are evaluated in isolation rather than jointly optimized. While we do not run psychophysical validation here, we show GSTN remains compatible with spatially-pooled synthesis pipelines.


\begin{figure}[t]
  \centering
  \begin{minipage}[c]{0.65\textwidth}
    \centering
    \includegraphics[width=\linewidth]{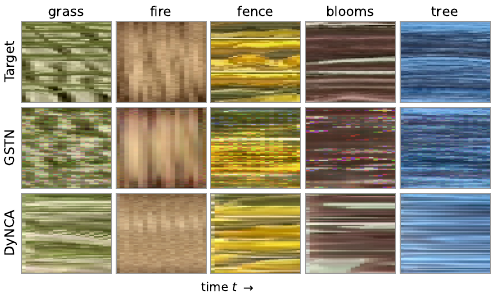}
  \end{minipage}\hfill
  \begin{minipage}[c]{0.35 \textwidth}
    \centering
    \scriptsize
    \setlength{\tabcolsep}{4pt}
    \renewcommand{\arraystretch}{1.1}
    \begin{tabular}{@{}lrr@{}}
      \toprule
      Metric & GSTN & DyNCA \\
      \midrule
      \multicolumn{3}{@{}l}{\textit{Deep-feature appearance}} \\
      LPIPS-Alex $\downarrow$ & 0.323 & \textbf{0.159} \\
      LPIPS-VGG $\downarrow$  & 0.515 & \textbf{0.323} \\
      DISTS $\downarrow$      & 0.324 & \textbf{0.199} \\
      FID $\downarrow$        & 162   & \textbf{83} \\
      CMMD $\downarrow$       & 2.441 & \textbf{1.339} \\
      \midrule
      \multicolumn{3}{@{}l}{\textit{Pixel-level}} \\
      SSIM $\uparrow$         & \textbf{0.114} & 0.107 \\
      MSE $\downarrow$        & \textbf{0.044} & \textbf{0.044} \\
      PSNR $\uparrow$         & \textbf{14.2}  & 14.1 \\
      \midrule
      \multicolumn{3}{@{}l}{\textit{Temporal}} \\
      Spectrum dist.\ $\downarrow$       & \textbf{0.176} & 0.518 \\
      $|\Delta$Curvature$|$ $\downarrow$ & \textbf{12.5}  & 17.8 \\
      $|\Delta$Diversity$|$ $\downarrow$ & 0.065 & \textbf{0.037} \\
      \bottomrule
    \end{tabular}
  \end{minipage}
  \caption{Video texture synthesis. Five DynTex \cite{dyntex} videos chosen to span the space of dynamic texture types. \textbf{Left:} space--time ($x$--$t$) slices: each panel shows one row of video pixels (vertical) across all frames, across time (horizontal). GSTN synths motion is similar to targets,  with less fine spatial detail and some chromatic noise. DyNCA under-represents motion with superior spatial matching. \textbf{Right:} mean over all five Dyntex videos (per-video results in Appendix Table~\ref{tab:video_synthesis_quality_quant}). DyNCA leads on deep-feature appearance metrics~\citep{heusel2017gans, jayasumana2024} and diversity match; GSTN leads on temporal power-spectrum match and perceptual straightness~\citep{henaff2019perceptual}. See supplemental material for videos.} 
  \label{fig:videosynths}
\end{figure}

\textbf{Dynamic Textures:} Extending human vision inspired models such as the Portilla--Simoncelli image model~\citep{portilla2000parametric} to the spatiotemporal domain is non-trivial, as its hand-selected cross-scale and cross-orientation statistics lack clear analogues in the time domain. Instead of hand picking important statistics, our method, leveraging all combinations of filter correlation statistics with the Gram matrix, can be naturally extended to video textures. We extend the spatial pyramid into the time domain by designing a bank of 3D Gabor filters arranged in stacked concentric rings in spatiotemporal frequency space, similar to oriented spacetime-energy representations ~\citep{derpanis2010dynamic,derpanis2012spacetime} with each filter paired with its quadrature complement. The filter bank is parameterized to span the spatiotemporal frequency range of human visual sensitivity \citep{kelly1979motion}; details and visualizations in Appendix.

We apply the same Gram matrix model to calculate correlations between our spatiotemporal filters, synthesizing texture videos by matching the set of filter correlation statistics. While contrastive learning in the spatiotemporal case is not computationally feasible given resource constraints, we are able to reduce our model size somewhat based on our findings from the spatial-only case, removing lowpass filters. Our model synthesizes videos that visually match well to the target textures (Figure \ref{fig:videosynths}), and results span a set of texture videos chosen to maximize variety in structure and motion properties, while maintaining computational tractability (Table \ref{tab:video_synthesis_quality_quant}). Quantitatively, cellular automaton model DyNCA ~\citep{pajouheshgar2023dynca} better matches spatial features, while GSTN better matches motion; videos are at \url{https://vdutell.github.io/gstn}. We do not claim improvement over prior dynamic-texture synthesis methods including \citep{tesfaldet2018two, pajouheshgar2023dynca}, but instead show that the same pyramid + correlation framework that succeeds on static textures extends naturally to video.

\section{Discussion}

GramStatTexNet merges the interpretability of biologically inspired multi-scale pyramid filters with the flexibility of Gramian-based correlations. By organizing statistics into structured filter families, we identify a compressed subset of correlations that retain synthesis quality while substantially reducing representation size, and expose which spatial-frequency, orientation, and color correlations carry texture identity. The same framework extends naturally to peripheral and dynamic textures via spatial pooling and a spatiotemporal filter bank, providing a unified, interpretable framework for modeling visual information across space, time, and eccentricity.

\textbf{Limitations} Our depletion analysis uses GSTN-wrap for computational efficiency; replication on GSTN-zpad is left to future work, though we expect qualitatively similar findings, with improvement gains mirroring the full statistics set. We provide no new neural or psychophysical validation of this representation, for static, peripheral, or dynamic textures; validation against human observers remains future work. Finally, contrastive compression is currently tractable only for the spatial model.

\textbf{Future work} Psychophysical evaluation of peripheral and dynamic-texture syntheses would test whether the compressed representation preserves perceptually relevant information. Scaling contrastive compression to spatiotemporal statistics would reveal which spatiotemporal correlations carry the most information for dynamic texture identity. Finally, the GSTN statistics could be matched via gradient-guidance with diffusion-based generative models, which we leave to future work.

\textbf{Broader Impacts} This is foundational research on interpretable, biologically-inspired visual representations, with no direct path to negative societal applications. Our syntheses are visibly lower-fidelity than current diffusion-based methods, so we do not meaningfully advance generative misuse with deepfakes. Potential positive impacts include: (i) interpretable, family-decomposed representations that link learned features to well-characterized properties of biological vision: our pyramid filters approximate V1 response properties, and the family taxonomy offers a hypothesis space, not validated here, for how V2 might combine V1 outputs, complementing macaque single-unit~\citep{ziemba2016selectivity} and macaque/human~\citep{freeman2013functional} evidence of V2 selectivity for the structured-family statistics; (ii) compressed representations that reduce storage and compute for downstream pipelines; and (iii) compatibility with foveated rendering, with potential applications in low-bandwidth AR/VR and accessibility tools.
\newpage
\section{Acknowledgments and Disclosures}
We thank the many individuals whose feedback and helpful discussions improved this paper, including Bruno Olshausen, Ayush Tewari, Prafull Sharma, Eric Li, Shaden Alshammari, as well as the anonymous reviewers and AC. This work was funded by the CSAIL MEnTorEd Opportunities in Research (METEOR) Fellowship, US National Science Foundation under grant number 1955219, as well as National Science Foundation Grant BCS-1826757 to PI Rosenholtz. The authors acknowledge the MIT SuperCloud \citep{reuther2018interactive} and Lincoln Laboratory Supercomputing Center for providing HPC resources that have contributed to the research results reported within
this paper.


{
    \small
    \bibliographystyle{ieeenat_fullname}
    \bibliography{main}
}

\clearpage
\appendix

\section{Appendix}
\FloatBarrier
\subsection{Image and Video Synthesis}
We synthesized visualized textures following the synthesis procedure in \citet{koevesdi2023stattexnet}, using the filter portion of the codebase from \citep{brown2021efficient}, with the pyramid settings at 4 orientations, 6 edge levels, edge start at level 1, with marginal statistics included. For image synthesis batched over the DTD dataset, we used Prodigy optimizer \citep{mishchenko2023prodigy}. For image synthesis on individual textures, foveated synthesis, and spatiotemporal generation, we used the L-BFGS solver \citep{liu1989limited} as demonstrated for textures in \citep{gatys2015texture}. See Appendix section \ref{sec:optimizers} for details on optimizers.

\paragraph{Synthesis Baselines} Syntheses for random filters \citep{ustyuzhaninov2017does} were created by a custom Pytorch port from the official Theano repo (https://github.com/ivust/random-texture-synthesis) with 20k iterations and the Prodigy solver \citep{mishchenko2023prodigy}. Synths for Gatys et al \citep{gatys2015texture} were created from an unofficial Pytorch repo for \citep{gatys2015texture} (https://github.com/trsvchn/deep-textures). Synths for Portilla \& Simoncelli~\citep{portilla2000parametric} were created using the official color extension package in Matlab (https://www.cns.nyu.edu/pub/eero/colorTextureSynth.tgz). Syntheses for µNCA \citep{mordvintsev2021mu} used the canonical JAX notebook from the official repository (https://github.com/google-research/self-organising-systems), trained per texture at 256×256 (one seed; the polka-dotted failure was confirmed over three seeds). Table 3 reports this run; Figure 2 shows the 128×128 native-resolution run. DyNCA \citep{pajouheshgar2023dynca} syntheses used the authors' official implementation (https://github.com/IVRL/DyNCA), fitting both motion and appearance to the target video at 64×64 for 5,000 iterations, matched to the GSTN video resolution.

\paragraph{GSTN-wrap vs.\ GSTN-zpad.}\label{sec:wrapzpad} GSTN-wrap and GSTN-zpad differ in the boundary condition of the steerable-pyramid convolutions. GSTN-wrap assumes the texture is stationary and tile-able and is faster, but does not retain edge boundaries. GSTN-zpad instead pads the image with a border of zeros before the Fourier transform, so opposite edges no longer interact and the boundary behaves as a non-repeating edge with no tiling assumption, at the cost of compute time.

\paragraph{Peripheral Synthesis} Spatial pooling of our model for peripheral syntheses was incorporated by tiling the image with overlapping pooling regions of 16x16 pixels tiled across the image, then synthesizing our texture model at each region in parallel. We utilize the pooling method from \citep{brown2021efficient}, warping the image, synthesizing over uniformly-tiled regions in the warped image space, and un-warping after synthesis, with parameters matched as closely as possible to \citep{freeman2011metamers}. 

\paragraph{Video Synthesis}To synthesize novel videos matching the statistics of input videos, we follow the same procedure as for images, calculating the spatiotemporal statistics vector from a target video. Then, starting with a white-noise video, we use L-BFGS to adjust pixel values in the synthetic video, using backpropagation of the mean squared error in statistics space between the synthetic and target video. This results in a synthetic video that matches the statistics of the original target video, but with scrambling of absolute spatial and temporal localization. Given this loss of spatial and temporal information, we use texture videos from the DynTex Database \citep{dyntex}, which we spatially center-crop and downsample to 64x64 to minimize artifacts and ensure uniformity using 20 frames at 15 fps (DynTex IDs 54pb110, 54pi110, 55fc210, 6440110, 6441210).

\FloatBarrier
\subsection{Optimizers: L-BFGS, Madgrad, Prodigy Comparison}
\label{sec:optimizers}
\begin{figure}[H]
  \centering
  \includegraphics[width=0.98\textwidth]{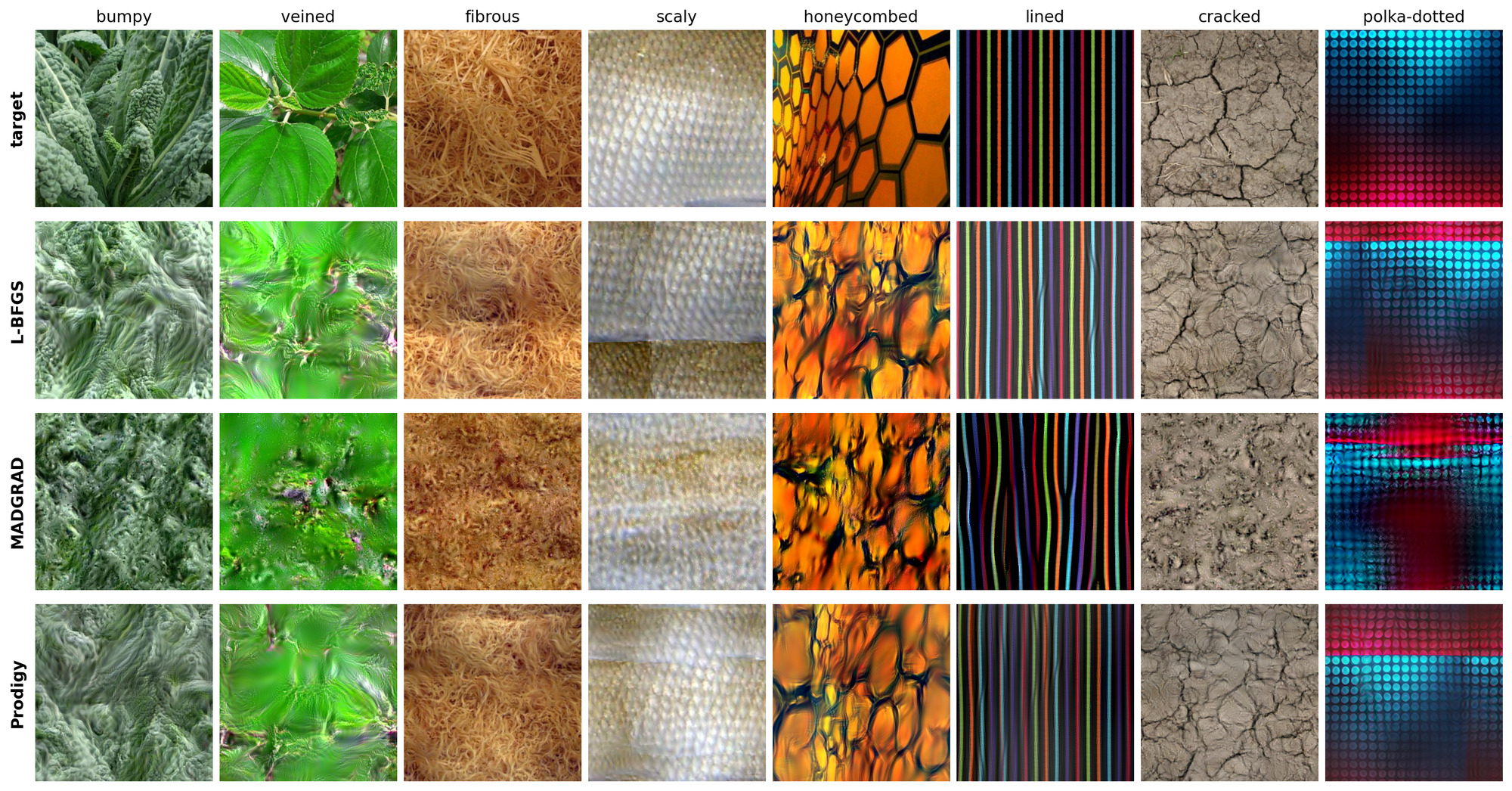}
  \caption{We find that Prodigy and Madgrad produce good quality synthesis like L-BFGS, and are batch compatible, saving significant time when synthesizing at scale, with Prodigy overall achieving slightly higher quality than Madgrad.}
  \label{fig:lbfgsoptimizer}
\end{figure}

L-BFGS \citep{liu1989limited} is often the optimizer of choice for statistical texture synthesis given its speed and efficiency at jointly optimizing the pixel space to find the target image, due to it being a second order method. However, the calculation of the Hessian makes batching image syntheses problematic, and non-batched synthesis on dataset scale intractably slow. To address this, we explored the use of MADGRAD \citep{defazio2021adaptivity} and Prodigy \citep{mishchenko2023prodigy} as alternative optimizers that are batch-compatible, and produce synthesis comparable to L-BFGS, the de-facto gold standard for pixel-backpropagation based synthesis (Figure \ref{fig:lbfgsoptimizer}). L-BFGS uses 100 outer iterations of Strong-Wolfe-line-searched LBFGS (lr=0.5, max\_iter=100, history=20, tolerance\_grad=1e-7); MADGRAD uses 20k iterations at lr=0.07, momentum=0.99; Prodigy uses 20k iterations with auto-tuned step size (lr=0.04, d0=1e-2, betas=(0.99, 0.999), safeguard\_warmup=True). All optimizers minimize MSE between target and synthesis Gram-statistic vectors with wrap or zero-padded mode, and the best-loss iterate is reported.

The choice of optimizer was determined by individual experiment requirements. For single-image qualitative comparisons, foveated peripheral synthesis, and video synthesis, we use L-BFGS, which converges efficiently on individual targets due to its second-order Hessian approximation. However, as discussed above, the Hessian computation does not separate cleanly across batched images, making L-BFGS intractable for large-scale evaluation. For experiments requiring batched synthesis across the validation set including texture variety, depletion experiments, and all quantitative results, we use Prodigy.

\FloatBarrier
\subsection{Computational Resources}
For the image synthesis experiments, synthesizing a single condition for the DTD-141 (141 subset) used two NVIDIA GeForce RTX 4090 GPUs, taking approximately 3-4 hours for the GSTN-wrap, and 6-8 hours for the GSTN-zpad condition with the Prodigy solver. Total estimated runtime for the image experiments is 1-2 weeks GPU time. For the video syntheses, synthesis ran on a single NVIDIA A-100 and took 1-2 hours for a single video. Foveated syntheses ran on a single NVIDIA A-100 and took 2-4 hours. Storage was <100GB for all datasets and experiments. Many additional preliminary and failed experiments, testing different optimizers, hyperparameters, resolutions, and pyramid settings were conducted than were reported in the paper, and while none individually used significant resources, altogether the use was not negligible.

\FloatBarrier
\subsection{Datasets}
For the spatial texture model, we utilized the Describable Textures Database (DTD) \citep{cimpoi2014describing}, keeping the largest centered square and resizing the image to 256x256 for synthesis, and 5-crops for the contrastive learning. For training the contrastive learning module, we used the entire training subset of split 1, which samples all 47 texture types. For the evaluations of our syntheses and depletions, for computational tractability, we sampled 3 randomly chosen images from each category of the validation subset of validation split 1, for a total of 141 images (DTD-141). For computational reasons, we also evaluate \citep{portilla2000parametric}, \citep{gatys2015texture}, and \citep{ustyuzhaninov2017does} on this DTD-141 subset. Because \citep{mordvintsev2021mu} takes many hours for a single synthesis, it was evaluated only on the 11-texture subset of Figure \ref{fig:imsynths}.

For the video texture model, we utilized the Dynamic Texture Database (DynTex) \citep{dyntex}. For computational tractability, we synthesized a small subset of videos selected to vary in frequency, shape, and motion type, but do not evaluate on the entire dataset.

\FloatBarrier
\subsection{Full Statistics Image Synthesis}

\begin{figure}[H]
  \centering
  \includegraphics[width=\textwidth]{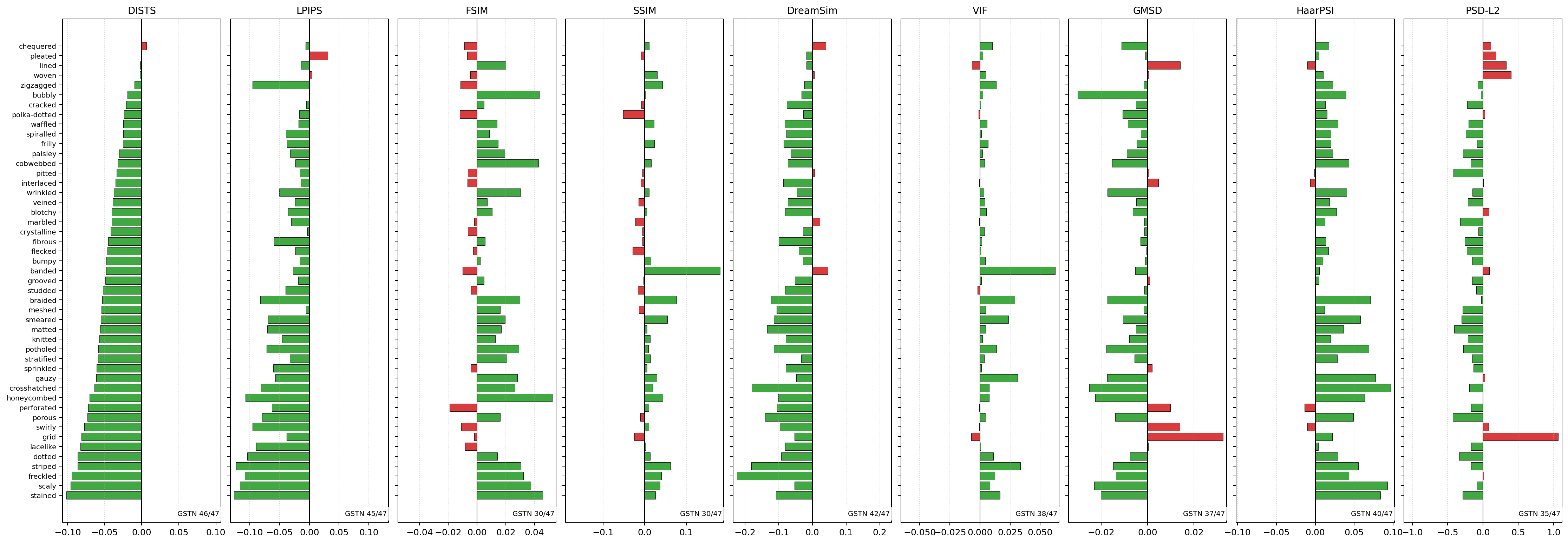}
  \caption{Per-category quantitative comparison of GSTN-zpad vs P\&S~\citep{portilla2000parametric} across nine perceptual and structural metrics. Each bar shows the mean per-category metric difference (GSTN $-$ P\&S); green = GSTN wins, red = P\&S wins. Categories sorted by DISTS-difference. Metrics span deep-network features (DISTS, LPIPS, DreamSim), local moments (FSIM, SSIM), and four bias-neutral metrics independent of either family (VIF, GMSD, HaarPSI, PSD-L2). DTD-141, 47 DTD categories with n=3 per category. GSTN-zpad wins the majority of categories on every metric (per-panel counts: DISTS 46/47, LPIPS 45/47, DreamSim 42/47, FSIM 30/47, SSIM 30/47, VIF 38/47, GMSD 37/47, HaarPSI 40/47, PSD-L2 35/47).}
  \label{fig:gstn_v_ps}
\end{figure}

\begin{figure}[H]
  \centering
  \includegraphics[width=\textwidth]{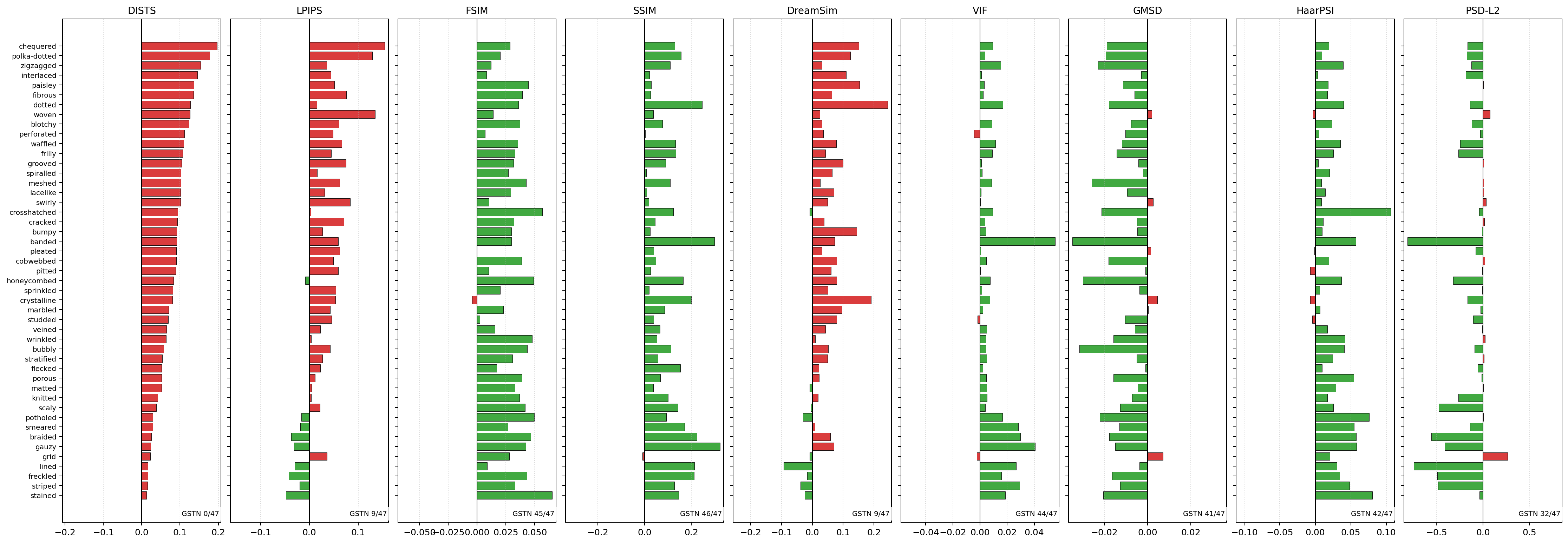}
  \caption{Per-category quantitative comparison of GSTN-zpad vs Gatys ~\citep{gatys2015texture}, same layout as Figure \ref{fig:gstn_v_ps}. DTD-141, 47 DTD categories with n=3 per category. Gatys wins on deep-network feature metrics (GSTN wins only DISTS 0/47, LPIPS 9/47, DreamSim 9/47) the metric family closest to Gatys' own VGG-feature training objective, and deep-learning based DreamSim, while GSTN-zpad wins on local-moment metrics (FSIM 45/47, SSIM 46/47) and on all four bias-neutral metrics (VIF 44/47, GMSD 41/47, HaarPSI 42/47, PSD-L2 32/47), despite using 30× fewer parameters (29K vs 852K).}
  \label{fig:gstn_v_gatys}
\end{figure}

We compare quantitative performance of GSTN per texture category over DTD-141 on a variety of metrics. A set of bias-neutral metrics (VIF, GMSD, HaarPSI, PSD-L2) was added to address the concern that local-moment metrics (SSIM, FSIM) are structurally aligned with GSTN's Gram-of-pyramid-bands representation while deep-feature metrics (DISTS~\citep{ding2020image}, LPIPS~\citep{zhang2018unreasonable}, DreamSim~\citep{fu2023dreamsim}) are aligned with Gatys' deep-VGG training objective. VIF ~\citep{sheikh2006image} is grounded in natural-scene statistics and Gaussian scale-mixture wavelet models; GMSD  \citep{xue2013gradient} compares the spatial deviation of local gradient-magnitude similarity; HaarPSI \citep{reisenhofer2018haar} uses Haar-wavelet feature similarity; PSD-L2 is the L2 distance between log power-spectral densities. None of these share architecture or training objective with GSTN or Gatys, but all four consistently favor GSTN-zpad in per-category comparisons (Figures \ref{fig:gstn_v_ps}, \ref{fig:gstn_v_gatys}), supporting that GSTN-zpad's advantage over P\&S is not a metric-bias artifact; against Gatys, results split by metric family (Figure~\ref{fig:gstn_v_gatys}).

\FloatBarrier
\subsection{Seed Sensitivity}

\begin{figure}[p]
\centering
\includegraphics[height=0.92\textheight]{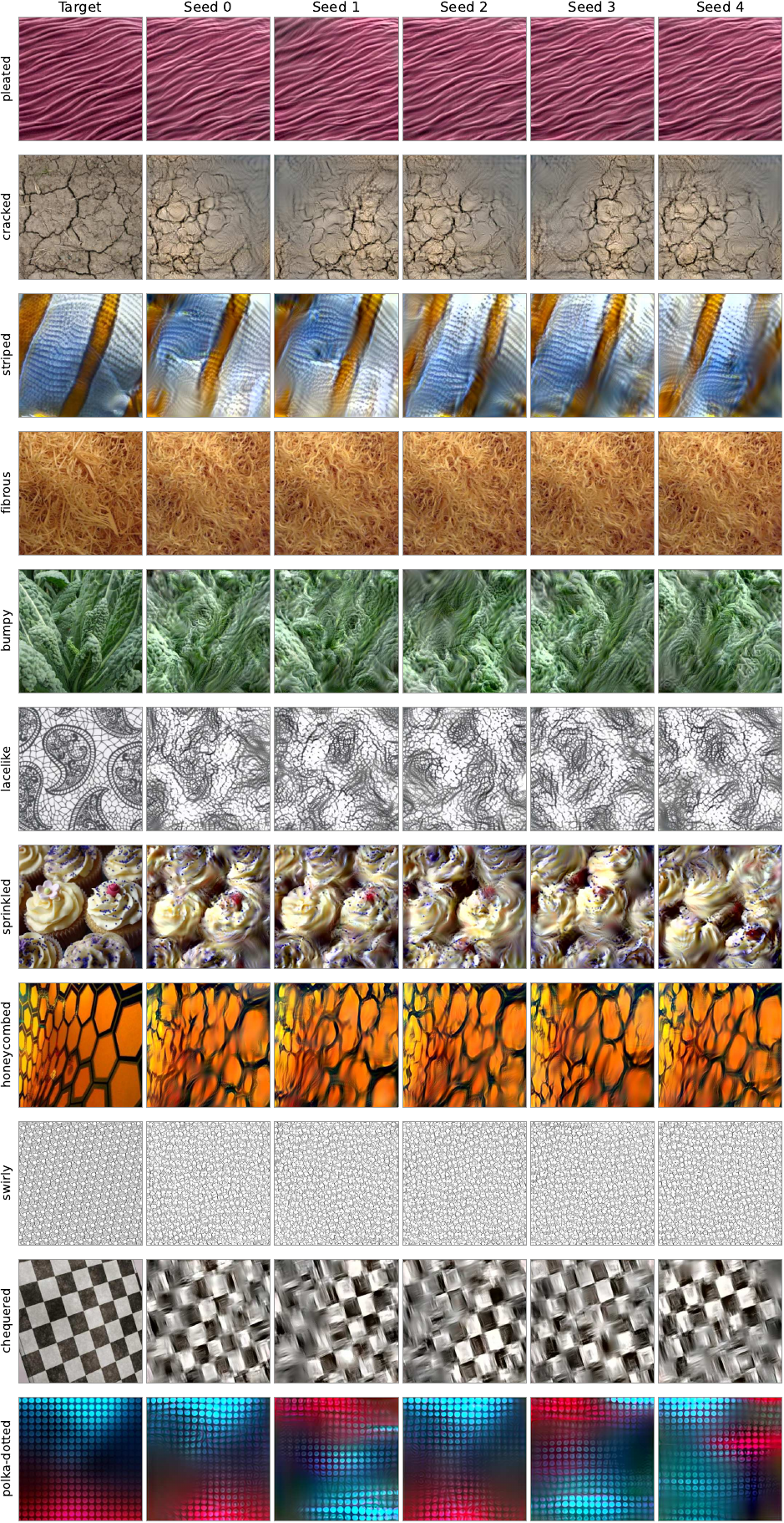}
\caption{\textbf{Synthesis is stable across random initializations.} The 11 textures of Figure~\ref{fig:imsynths}, each synthesized by GSTN-zpad from five independent random initializations, shown against the target.}
\label{fig:seed_grid}
\end{figure}

We tested the sensitivity of our method to multiple seeds to evaluate the stability of synthesis quality (Table \ref{tab:seeds}, Figure \ref{fig:seed_grid}). We find that both qualitatively and quantitatively, texture identity accounts for far more of the variation than seed, and this trend is consistent across all metrics tested.
\begin{table}[h]
\centering
\caption{\textbf{Sensitivity to random initialization.} The 11 textures of
Figure~\ref{fig:imsynths}, each synthesized from 5 independent random initializations with GSTN-zpad. Entries are mean $\pm$ s.d.\ across seeds.}
\label{tab:seeds}
\resizebox{\textwidth}{!}{%
\begin{tabular}{lrrrrrrrrr}
\toprule
Texture & DISTS $\downarrow$ & LPIPS $\downarrow$ & DrmSim $\downarrow$ & FSIM $\uparrow$
 & SSIM $\uparrow$ & VIF $\uparrow$ & GMSD $\downarrow$ & HaarPSI $\uparrow$ & PSD-L2 $\downarrow$ \\
\midrule
\texttt{pleated\_0136}     & 0.231 $\pm$ 0.005 & 0.332 $\pm$ 0.005 & 0.086 $\pm$ 0.018 & 0.639 $\pm$ 0.004 & 0.078 $\pm$ 0.019 & 0.032 $\pm$ 0.003 & 0.263 $\pm$ 0.001 & 0.434 $\pm$ 0.003 & 1.481 $\pm$ 0.009 \\
\texttt{cracked\_0056}     & 0.243 $\pm$ 0.008 & 0.587 $\pm$ 0.010 & 0.317 $\pm$ 0.024 & 0.615 $\pm$ 0.005 & 0.098 $\pm$ 0.005 & 0.007 $\pm$ 0.001 & 0.251 $\pm$ 0.004 & 0.374 $\pm$ 0.008 & 1.740 $\pm$ 0.006 \\
\texttt{striped\_0050}     & 0.246 $\pm$ 0.020 & 0.633 $\pm$ 0.099 & 0.215 $\pm$ 0.028 & 0.663 $\pm$ 0.032 & 0.171 $\pm$ 0.031 & 0.021 $\pm$ 0.005 & 0.240 $\pm$ 0.012 & 0.314 $\pm$ 0.033 & 1.386 $\pm$ 0.025 \\
\texttt{fibrous\_0112}     & 0.269 $\pm$ 0.008 & 0.524 $\pm$ 0.005 & 0.167 $\pm$ 0.002 & 0.698 $\pm$ 0.002 & 0.102 $\pm$ 0.008 & 0.009 $\pm$ 0.001 & 0.208 $\pm$ 0.001 & 0.439 $\pm$ 0.002 & 1.677 $\pm$ 0.003 \\
\texttt{bumpy\_0077}       & 0.283 $\pm$ 0.008 & 0.568 $\pm$ 0.012 & 0.282 $\pm$ 0.024 & 0.669 $\pm$ 0.010 & 0.061 $\pm$ 0.011 & 0.009 $\pm$ 0.002 & 0.231 $\pm$ 0.006 & 0.377 $\pm$ 0.013 & 1.709 $\pm$ 0.005 \\
\texttt{lacelike\_0004}    & 0.307 $\pm$ 0.005 & 0.513 $\pm$ 0.008 & 0.252 $\pm$ 0.009 & 0.620 $\pm$ 0.007 & 0.015 $\pm$ 0.008 & 0.005 $\pm$ 0.001 & 0.254 $\pm$ 0.004 & 0.382 $\pm$ 0.006 & 1.766 $\pm$ 0.002 \\
\texttt{sprinkled\_0042}   & 0.310 $\pm$ 0.012 & 0.667 $\pm$ 0.031 & 0.441 $\pm$ 0.022 & 0.601 $\pm$ 0.015 & 0.085 $\pm$ 0.027 & 0.017 $\pm$ 0.008 & 0.266 $\pm$ 0.008 & 0.288 $\pm$ 0.020 & 1.675 $\pm$ 0.011 \\
\texttt{honeycombed\_0149} & 0.329 $\pm$ 0.008 & 0.647 $\pm$ 0.023 & 0.287 $\pm$ 0.007 & 0.554 $\pm$ 0.012 & 0.171 $\pm$ 0.013 & 0.026 $\pm$ 0.006 & 0.304 $\pm$ 0.006 & 0.244 $\pm$ 0.017 & 1.664 $\pm$ 0.009 \\
\texttt{swirly\_0065}      & 0.344 $\pm$ 0.004 & 0.475 $\pm$ 0.003 & 0.111 $\pm$ 0.006 & 0.605 $\pm$ 0.002 & 0.010 $\pm$ 0.006 & 0.003 $\pm$ 0.000 & 0.218 $\pm$ 0.001 & 0.436 $\pm$ 0.002 & 3.347 $\pm$ 0.007 \\
\texttt{chequered\_0177}   & 0.403 $\pm$ 0.011 & 0.682 $\pm$ 0.006 & 0.409 $\pm$ 0.008 & 0.567 $\pm$ 0.004 & 0.106 $\pm$ 0.016 & 0.014 $\pm$ 0.005 & 0.297 $\pm$ 0.002 & 0.252 $\pm$ 0.005 & 1.748 $\pm$ 0.019 \\
\texttt{polka-dotted\_0140}& 0.454 $\pm$ 0.053 & 0.609 $\pm$ 0.119 & 0.144 $\pm$ 0.026 & 0.546 $\pm$ 0.041 & 0.169 $\pm$ 0.059 & 0.017 $\pm$ 0.013 & 0.238 $\pm$ 0.023 & 0.317 $\pm$ 0.066 & 2.456 $\pm$ 0.018 \\
\bottomrule
\end{tabular}}
\end{table}

\FloatBarrier
\subsection{Quantitative Comparisons on 11-Texture Subset}

Table~\ref{tab:munca} reports quantitative performance on the 11 textures of Figure~\ref{fig:imsynths}, including $\mu$NCA. GSTN is comparable to Gatys and Ustyuzhaninov on many metrics at a much smaller size, while $\mu$NCA, the smallest model, is worst on 7 of 9 metrics.

\begin{table}[h]
\centering
\scriptsize
\setlength{\tabcolsep}{2pt}
\caption{Comparative synthesis quality for the 11 textures of Figure~\ref{fig:imsynths}. Values in parentheses exclude polka-dotted, on which $\mu$NCA failed across 3 seeds. Bold = best per column, computed separately for each value (both when tied). $\mu$NCA's per-texture training (1--2\,h per image on an RTX 4090) made evaluation on DTD-141 intractable.}
\label{tab:munca}
\begin{tabular}{@{}lrrrrrrrrrr@{}}
\toprule
Method & \# Stats & DISTS\,$\downarrow$ & LPIPS\,$\downarrow$ & DrmSim\,$\downarrow$ & FSIM\,$\uparrow$ & SSIM\,$\uparrow$ & VIF\,$\uparrow$ & GMSD\,$\downarrow$ & HaarPSI\,$\uparrow$ & PSD-L2\,$\downarrow$ \\
\midrule
GSTN-zpad & 28,929 & 0.29 (0.28) & 0.53 (0.54) & 0.22 (0.23) & \textbf{0.63} (\textbf{0.64}) & 0.12 (0.10) & 0.02 (\textbf{0.02}) & \textbf{0.24} (\textbf{0.25}) & \textbf{0.38} (\textbf{0.37}) & 1.88 (1.81) \\
GSTN-wrap & 28,929 & 0.34 (0.32) & 0.60 (0.60) & 0.26 (0.27) & 0.61 (0.61) & 0.08 (0.07) & 0.01 (0.01) & 0.25 (0.26) & 0.34 (0.34) & 1.95 (1.87) \\
P\&S~\citep{portilla2000parametric} & $\sim$8,400 & 0.36 (0.35) & 0.63 (0.63) & 0.30 (0.31) & 0.62 (0.62) & 0.08 (0.07) & 0.01 (0.01) & 0.26 (0.26) & 0.34 (0.35) & 1.83 (1.82) \\
Ustyuzh.~\citep{ustyuzhaninov2017does} & 524,800 & 0.25 (0.25) & \textbf{0.49} (0.52) & \textbf{0.15} (\textbf{0.16}) & \textbf{0.63} (0.62) & \textbf{0.15} (0.10) & \textbf{0.03} (0.01) & 0.25 (0.26) & \textbf{0.38} (0.35) & \textbf{1.57} (\textbf{1.60}) \\
Gatys~\citep{gatys2015texture} & 852k & \textbf{0.21} (\textbf{0.20}) & 0.51 (\textbf{0.51}) & \textbf{0.15} (\textbf{0.16}) & 0.59 (0.60) & 0.05 (0.04) & 0.01 (0.01) & 0.26 (0.26) & 0.33 (0.34) & 2.05 (1.96) \\
$\mu$NCA~\citep{mordvintsev2021mu} & 588 & 0.41 (0.39) & 0.64 (0.63) & 0.39 (0.35) & 0.53 (0.55) & \textbf{0.15} (\textbf{0.14}) & 0.02 (\textbf{0.02}) & 0.30 (0.29) & 0.28 (0.30) & 3.00 (3.04) \\
\bottomrule
\end{tabular}
\end{table}

\FloatBarrier
\subsection{Efficiency Comparisons}

We performed an efficiency evaluation across models, size, inference, and synthesis time (Table ~\ref{tab:efficiency}). Overall, GSTN shows balanced efficiency compared to other methods, with the lowest GPU memory load, and with Prodigy optimizer can be highly parallelized on a single GPU to decrease effective per-image synthesis time. While Gatys dominates in speed, its representation size (number of statistics) is very large in comparison. And while $\mu$NCA is by far the smallest model in terms of number of statistics, as a cellular automaton it unrolls over many steps at full resolution, leading to a huge effective memory footprint, and its synthesis time is by far the worst (though the synthesis time may perform better and be parallelizable natively on a TPU).
\begin{table}[h]
\centering
\small
\setlength{\tabcolsep}{4pt}
\caption{\textbf{Efficiency Comparisons} All timings and memory are measured on a single idle NVIDIA RTX 4090 GPU (P\&S on Apple Macbook Air M2 with MATLAB as the model/codebase is not GPU compatible). Reduced statistics sets for GSTN and Gatys represent ablations reported that exhibit little loss of quality compared to the full models. Iteration counts differ by method, so per-image time reflects optimization budget as well as per-step cost. All inference and per-image synthesis times reported are non-batched. With Prodigy, GSTN syntheses can be batched (36 images on a single 4090), which reduces effective per-image cost substantially. Batching is limited by the optimizer rather than the model; MADGRAD or Prodigy could in principle replace L-BFGS in Gatys and Ustyuzhaninov, but here we report the standard configurations.}
\label{tab:efficiency}
\begin{tabular}{lrrrrl}
\toprule
Method & \shortstack[r]{Statistic size\\(reduced) $\downarrow$}
       & \shortstack[r]{Inference\\ms/eval $\downarrow$}
       & \shortstack[r]{Synthesis / image\\(non-parallel) $\downarrow$}
       & \shortstack[r]{Peak GPU\\memory $\downarrow$}
       & Parallelizable? \\
\midrule
GSTN-zpad (Prodigy)        & 28,929 (10k) & 13.9  & 4.6 min & 0.37 GB & \textbf{yes (GPU batch)} \\
GSTN-wrap (Prodigy)        & 28,929 (10k) & 13.3  & 4.4 min & \textbf{0.30} GB & \textbf{yes (GPU batch)} \\
GSTN-zpad (L-BFGS)         & 28,929 (10k) & 17.9  & 3.7 min & 0.53 GB & no (L-BFGS) \\
GSTN-wrap (L-BFGS)         & 28,929 (10k) & 16.5  & 2.2 min & 0.46 GB & no (L-BFGS) \\
\midrule
P\&S                       & 8k           & 700   & 34.8 s  & n/a (CPU) & \textbf{yes (MATLAB)} \\
Ustyuzhaninov              & 525k         & 114.4 & 8.3 min & 1.42 GB & no (L-BFGS) \\
Gatys (VGG-19)             & 852k (177k)  & \textbf{6.9}   & \textbf{6.9 s}  & 0.42 GB & no (L-BFGS) \\
$\mu$NCA ($256{\times}256$)   & \textbf{588}          & 686   & 91 min  & 7.73 GB & \textit{yes} (TPU only) \\
$\mu$NCA ($128{\times}128$, native) & \textbf{588}    & 526   & 70 min  & 1.98 GB & \textit{yes} (TPU only) \\
\bottomrule
\end{tabular}
\end{table}

\FloatBarrier
\subsection{Resolution Scaling}

\begin{figure}[h]
\centering
\includegraphics[width=\textwidth]{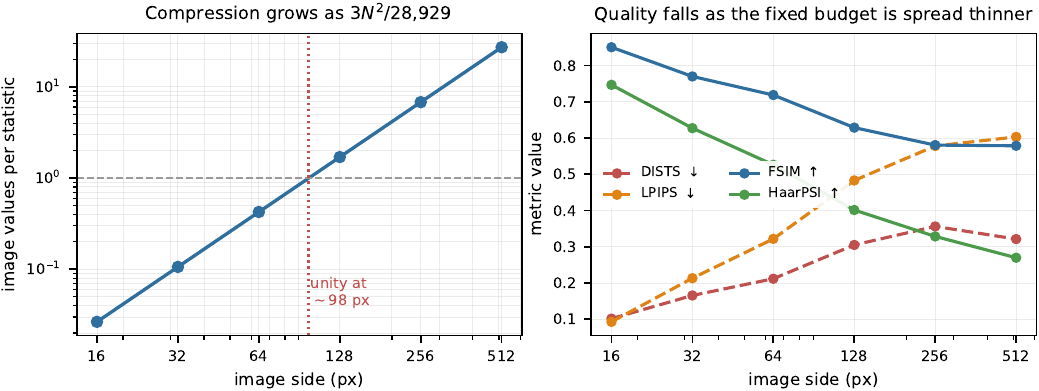}
\caption{\textbf{Compression and quality as a function of synthesis resolution.}
\textit{Left:} representation size is constant at 28{,}929 statistics, so compression grows as
$3N^2/28{,}929$ and GSTN becomes an under-complete representation above $\sim$98 px per side (dashed line) and over-complete below it. \textit{Right:} synthesis quality as a function of the synthesis resolution. FSIM, HaarPSI and LPIPS degrade monotonically as the budget is spread over larger images; DISTS peaks at $256{\times}256$ and improves slightly at $512{\times}512$. Below 64 px DISTS and LPIPS are outside their validated regime and are plotted for completeness.}
\label{fig:resolution_curves}
\end{figure}

\begin{figure}[p]
\centering
\includegraphics[width=0.75\textwidth]{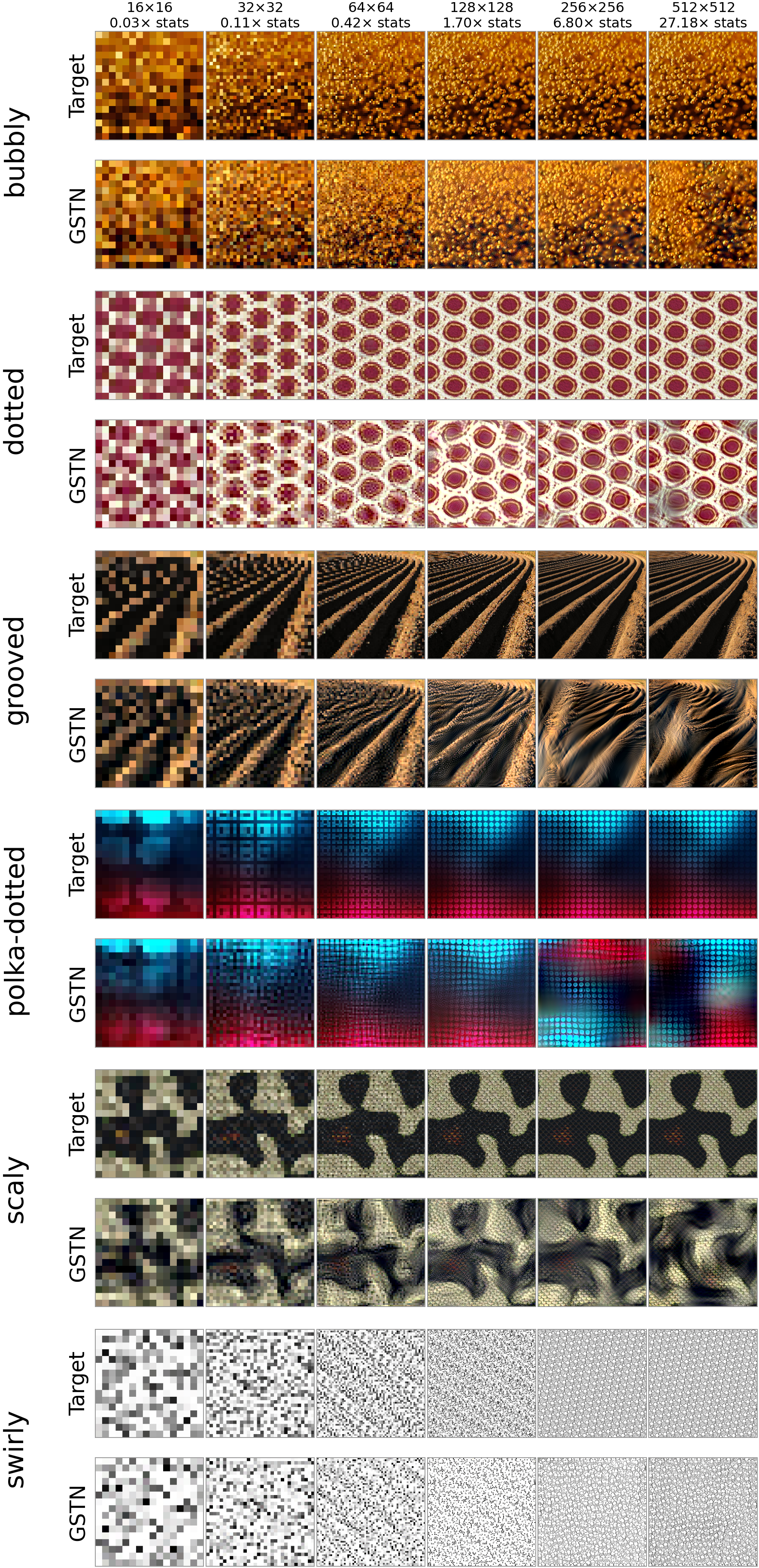}
\caption{\textbf{Synthesis quality with variable target resolution:} Because images of different target sizes are encoded with the same fixed representational budget, quality of match to target degrades as image size grows. Like synthesis quality generally, the magnitude of this degradation is texture dependent.}
\label{fig:resolution_grid}
\end{figure}

The full non-depleted GSTN model uses a fixed budget of 28,929 statistics to represent images, regardless of target image resolution (Figure \ref{fig:resolution_curves}), resulting in a varying amount of compression. Thus, the relative size efficiency of GSTN improves as image resolution increases, but at the tradeoff of synthesis quality. This effect is visible in the tiled syntheses of Figure \ref{fig:pooled_ablations} comparing the center (blue, smaller region) and edge (red, larger region) cutouts.  We tested the effect of differing image size on synthesis quality and wall clock given this fixed representational budget (Figure \ref{fig:resolution_grid}, Table ~\ref{tab:resolution}). We find that as expected, synthesis quality decreases with increased image size and increased ratio to representation size. This varying relationship mirrors the increase in pooling region size in peripheral syntheses and serves as a tractable but controlled and principled proxy for the eccentricity-dependent degradation shown in Figure~\ref{fig:pooled_ablations}. Likewise, synthesis time increases with target image size, although this relationship is sub-linear with respect to number of pixels, with a 512x512 pixel color image taking less than 7x the synthesis time of a 16x16 pixel color image.

\begin{table}[h]
\centering
\scriptsize
\setlength{\tabcolsep}{5pt}
\caption{\textbf{Scaling of compression, quality and wall-clock with image resolution.}
Six DTD textures (\texttt{bubbly\_0064}, \texttt{dotted\_0078}, \texttt{grooved\_0054},
\texttt{polka-dotted\_0140}, \texttt{scaly\_0168}, \texttt{swirly\_0065}) $\times$ 3 seeds
(18 syntheses per row), 20,000 iterations, GSTN-zpad.
Metrics are computed with synthesis and target both at the synthesis resolution. Representation
size is fixed at 28,929 statistics at every resolution, so compression (``values per
statistic'') grows with the square of the side length and crosses unity at $\sim\!98{\times}98$
px. Wall-clock scales strongly sublinearly: $1024\times$ the pixels costs $7\times$ the time.
Resolutions above $512{\times}512$ are untestable because DTD contains no $1024{\times}1024$
images. $^{*}$Below 64 px the VGG-based metrics (DISTS, LPIPS) operate outside their validated
regime, and VIF is undefined.}
\label{tab:resolution}
\begin{tabular}{lrrrrrrrrr}
\toprule
Resolution & \shortstack[r]{Image size\\($3N^2$)} & \shortstack[r]{Values\\per stat.}
 & DISTS $\downarrow$ & LPIPS $\downarrow$ & FSIM $\uparrow$ & GMSD $\downarrow$
 & HaarPSI $\uparrow$ & VIF $\uparrow$ & \shortstack[r]{Wall-clock\\/ synthesis} \\
\midrule
$16{\times}16^{*}$   & 768     & 0.027$\times$  & 0.101 & 0.093 & 0.851 & 0.101 & 0.747 & n/a   & 96 s \\
$32{\times}32^{*}$   & 3,072   & 0.106$\times$  & 0.165 & 0.213 & 0.770 & 0.136 & 0.627 & n/a   & 103 s \\
$64{\times}64$       & 12,288  & 0.425$\times$  & 0.212 & 0.322 & 0.719 & 0.199 & 0.526 & 0.049 & 154 s \\
$128{\times}128$     & 49,152  & 1.699$\times$  & 0.305 & 0.483 & 0.629 & 0.232 & 0.401 & 0.021 & 248 s \\
$256{\times}256$     & 196,608 & 6.796$\times$  & 0.356 & 0.578 & 0.581 & 0.253 & 0.328 & 0.015 & 313 s \\
$512{\times}512$     & 786,432 & 27.185$\times$ & 0.321 & 0.603 & 0.579 & 0.282 & 0.270 & 0.016 & 670 s \\
\bottomrule
\end{tabular}
\end{table}

\FloatBarrier
\subsection{Contrastive Learning}

For contrastive learning, we trained a single fully connected linear layer of size (28,929 x 100), reducing the size of our representation space by nearly 300 times. We used DTD train split 1, taking 5-crops (4 corners plus center) of the image of size 128x128 with random vertical and horizontal flips, and feeding these through the analysis pipeline, extracting the upper triangle of the Gramian matrix for 28,929 statistics, and training the network in batches of 50 textures. We trained the network for 100 epochs with the Normalized Temperature-Scaled Cross-Entropy (NT-Xent) loss from pytorch metric learning, defined as:

\begin{equation}
\mathcal{L}_i = -\log \frac{\exp(\mathrm{sim}(z_i, z_{j(i)}) / \tau)}{\sum_{k \neq i} \exp(\mathrm{sim}(z_i, z_k) / \tau)}
\end{equation}

\begin{figure}[H]
  \centering
  \includegraphics[width=0.98\textwidth]{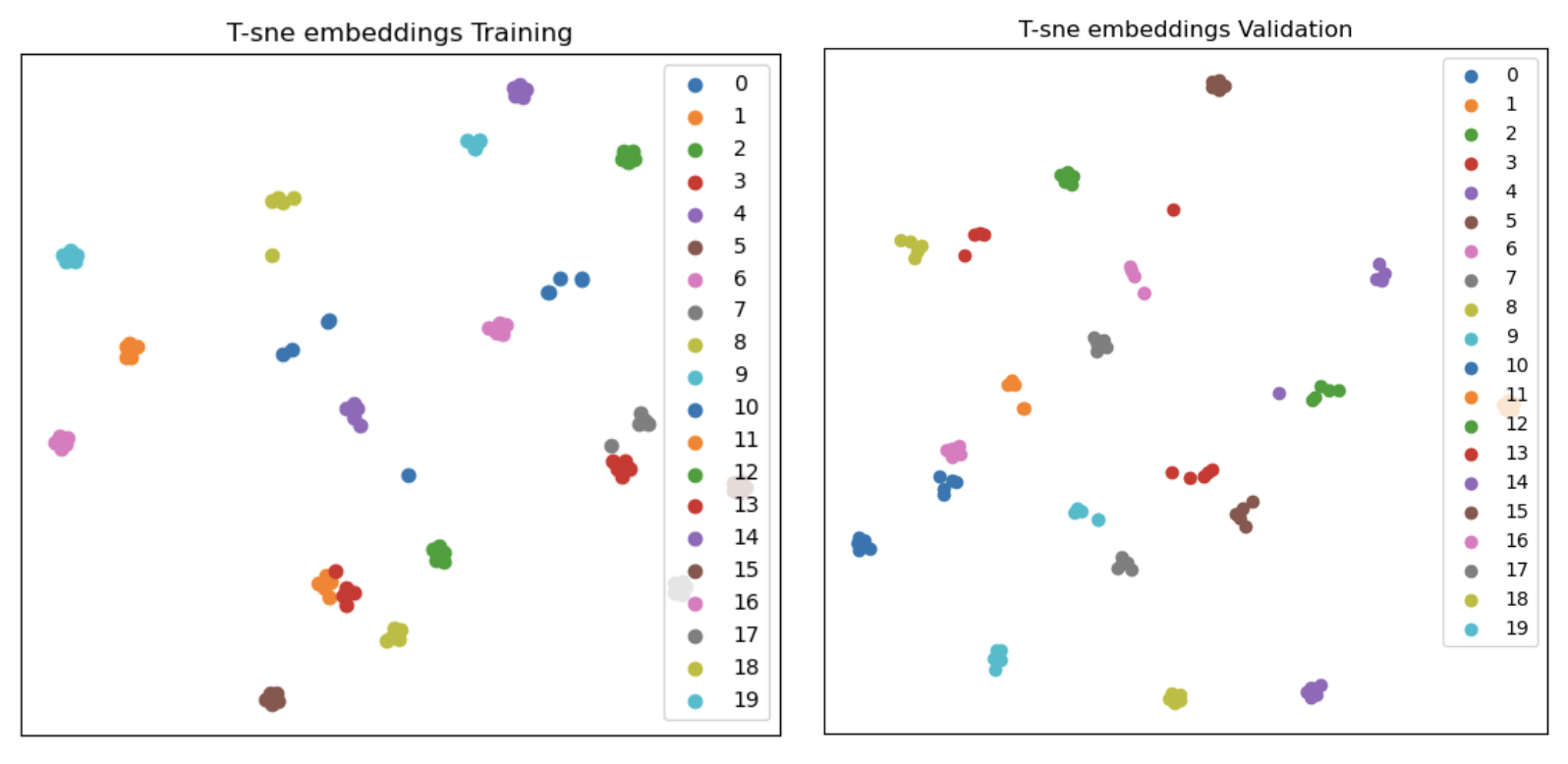}
  \caption{Contrastive learning successfully groups texture images cropped from the same parent image (same color), and separates images from different parent images. Training set (left). Validation set (right) }
  \label{fig:umap}
\end{figure}

Conceptually, this loss encourages the learned representation (100 dimensions) to represent textures from the same parent texture similarly, and textures from different parent textures differently. Figure~\ref{fig:umap} shows the learned embedding for training and validation textures. Crops drawn from the same parent texture cluster tightly in the 100-dimensional compressed space, while crops from different parent textures are well separated, both on training data and on the held-out validation split, indicating that the learned compression preserves texture identity rather than memorizing parent-image features.

\FloatBarrier
\subsection{Analytical Solution for Contrastive Learning}
\label{sec:cl_derivation}
Let $i$ denote the $i$th parent texture images, and $n(i)$ denotes the set of texture images cropped from $i$th parent texture images. Let $a(x)$ denote the texture statistic of image $x$, and let $P$ be the weight of the fully connected layer of the contrastive learning. The contrastive learning problem can be formulated as: 
\begin{align}
\min_{P} \sum_i \sum_{j,k \in n(i)} ||P a(x_j) - P a(x_k)||_F^2 \quad 
\end{align}
such that
\begin{align}
\quad P\mathbb{E}_x [a(x)a(x)^T] P^T = I    
\end{align}
This optimization problem can be written in the following matrix form: 

\begin{equation}
\min_P ||PAD||_F^2 \quad s.t. \quad PVP^T = I
\end{equation}
Where $V = \frac{1}{N}AA^T$, N is the total number of used images crops. $V$ is the covariance matrix of statistics. $A$ contains the statistic of all possible crops. Each column of $A$ is the $a(x_i)$ for some $i$. $D$ is analogous to the first order derivative operator defined on a graph. Specifically,   $D = [d_1, d_2, \cdots, d_M]$, where $M$ denotes the number of pair of crops in the training set. For $i$th pair of crops $x_k$ and $x_j$, we have $d_{ik} =1$ and $d_{ij}= -1$. All other entries of $D$ are zeros.

The solution to this generalized eigen-decomposition problem is given by $P = UV^{-\frac{1}{2}}$, where $U$ is a matrix of $L$ trailing eigenvectors (i.e. eigenvectors with the smallest eigenvalues) of the matrix $Q = V^{-\frac{1}{2}} A D D^T A^T V^{-\frac{1}{2}}$ \citep{chen2022minimalistic}. 

This means solving the contrastive learning problem requires calculating $Q$, which requires calculating the matrix $ADD^TA^T$ and $V^{-\frac{1}{2}}$. The latter can be easily computed by diagonalizing $V$. To compute $ADD^TA^T$ on the other hand is generally computational expensive. Luckily, since $D$ is highly sparse, we can efficiently calculate matrix $ADD^TA^T$ as the following: 
\begin{equation}
S =  2 \sum_i \left( |n(i)| \sum_{k \in n(i)} a_k a_k^T - \bar{a}_i \bar{a}_i^T \right)\end{equation}
where $\bar{a_i} = \sum_{j \in n(i)} a(x_j)$. Let $a_i$ denote $a(x_i)$ for convenience. The derivation is the following:

\begin{align*}
S &= A D D^T A^T \\
&= \sum_{i} \sum_{j,k \in n(i)} (a_k - a_j) (a_k - a_j)^T \\
&= \sum_{i} \sum_{j \in n(i)} \sum_{k \in n(i)} 
    a_k a_k^T + a_j a_j^T - a_k a_j^T - a_j a_k^T \\
&= \sum_{i} \sum_{j \in n(i)} ( \sum_{k \in n(i)} 
    a_k a_k^T + |n(i)| a_j a_j^T - \\
    &\quad(\sum_{k \in n(i)} a_k) a_j^T - a_j (\sum_{k \in n(i)} a_k)^T  ) \\
&= \sum_{i} ( |n(i)| \sum_{k \in n(i)} 
    a_k a_k^T + |n(i)| \sum_{j \in n(i)} a_j a_j^T -  \\
    &\quad(\sum_{k \in n(i)} a_k) (\sum_{j \in n(i)} a_j)^T - \\
    &\quad(\sum_{j \in n(i)} a_j) (\sum_{k \in n(i)} a_k)^T  ) \\    
&= \sum_{i} 2 |n(i)| \sum_{k \in n(i)} 
    a_k a_k^T  -  2  (\sum_{k \in n(i)} a_k) (\sum_{j \in n(i)} a_j)^T \\
&=  2 \sum_i \left( |n(i)| \sum_{k \in n(i)} a_k a_k^T - \bar{a}_i \bar{a}_i^T \right)
\end{align*}

\FloatBarrier
\subsection{Contribution of Statistics by Contrastive Learning}

\begin{figure}[H]
  \centering
  \includegraphics[width=0.98\textwidth]{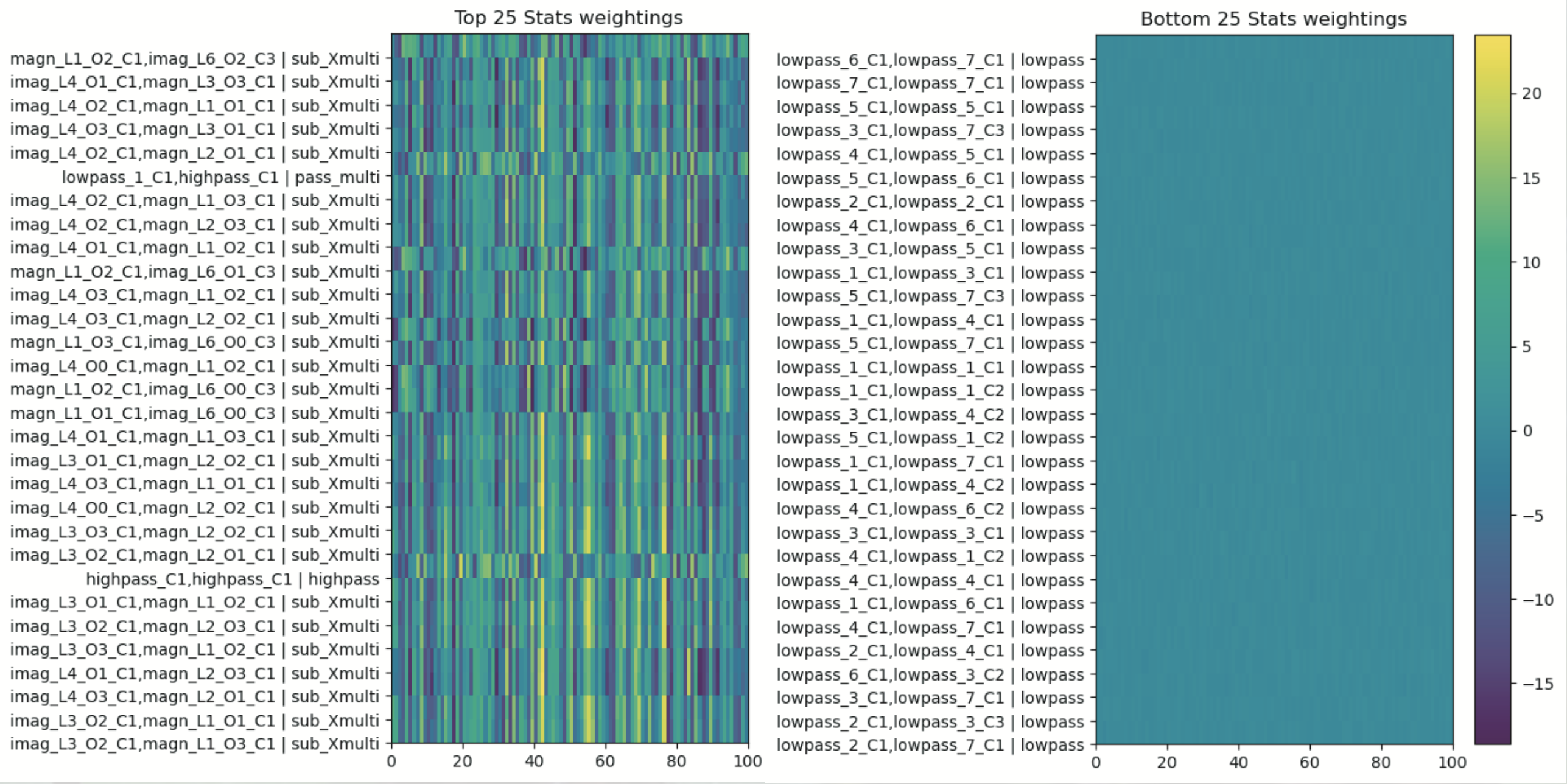}
  \caption{We use the weight matrix of the compression layer from the contrastive learning module to order correlation statistics by their contribution to the compressed representation. Learned weights of the single fully connected layer show high weightings for unstructured statistics, and lowest weightings for lowpass filter statistics.}
  \label{fig:weights}
\end{figure}

\begin{figure}[H]
  \centering
  \includegraphics[width=0.98\textwidth]{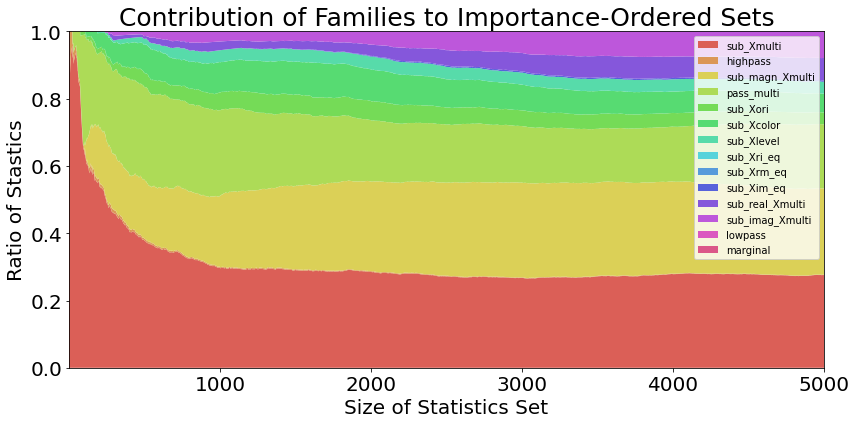}
  \caption{Using the contrastively learned importance ordering, we visualize the percent contribution of individual families to the total set, for varying statistics set size. Families Mixed-phase multi-attr. (sub\_Xmulti), highpass, and Non-oriented $\times$ band (pass\_multi) contribute most when selecting for the most important statistics (left).}
  \label{fig:familycontributions}
\end{figure}

Once the contrastive compression layer is trained, the weight matrix can be interpreted as an importance map over individual statistics: each column corresponds to one Gram-matrix statistic, and the magnitude of that column reflects how strongly the statistic contributes to the learned 100-dimensional representation. We compute a per-statistic importance score as the $L_1$ norm of the corresponding column, and rank all 28,929 statistics in descending order by this score. Figure~\ref{fig:weights} visualizes this ranking grouped by statistical family, showing that unstructured cross-correlation families (Mixed-phase multi-attr. (sub\_Xmulti) and Non-oriented $\times$ band (\texttt{pass\_multi})) receive the highest weights on average, while lowpass-filter statistics receive the lowest. Figure~\ref{fig:familycontributions} further breaks this down by reporting the percent contribution of each family to the top-N most important statistics for varying N, showing that the top-ranked statistics are dominated by a small number of large unstructured families.

\FloatBarrier
\subsection{Spatiotemporal Filters}
\label{sec:filtereqs}

\begin{figure}[h]
  \centering
  \includegraphics[width=0.95\textwidth]{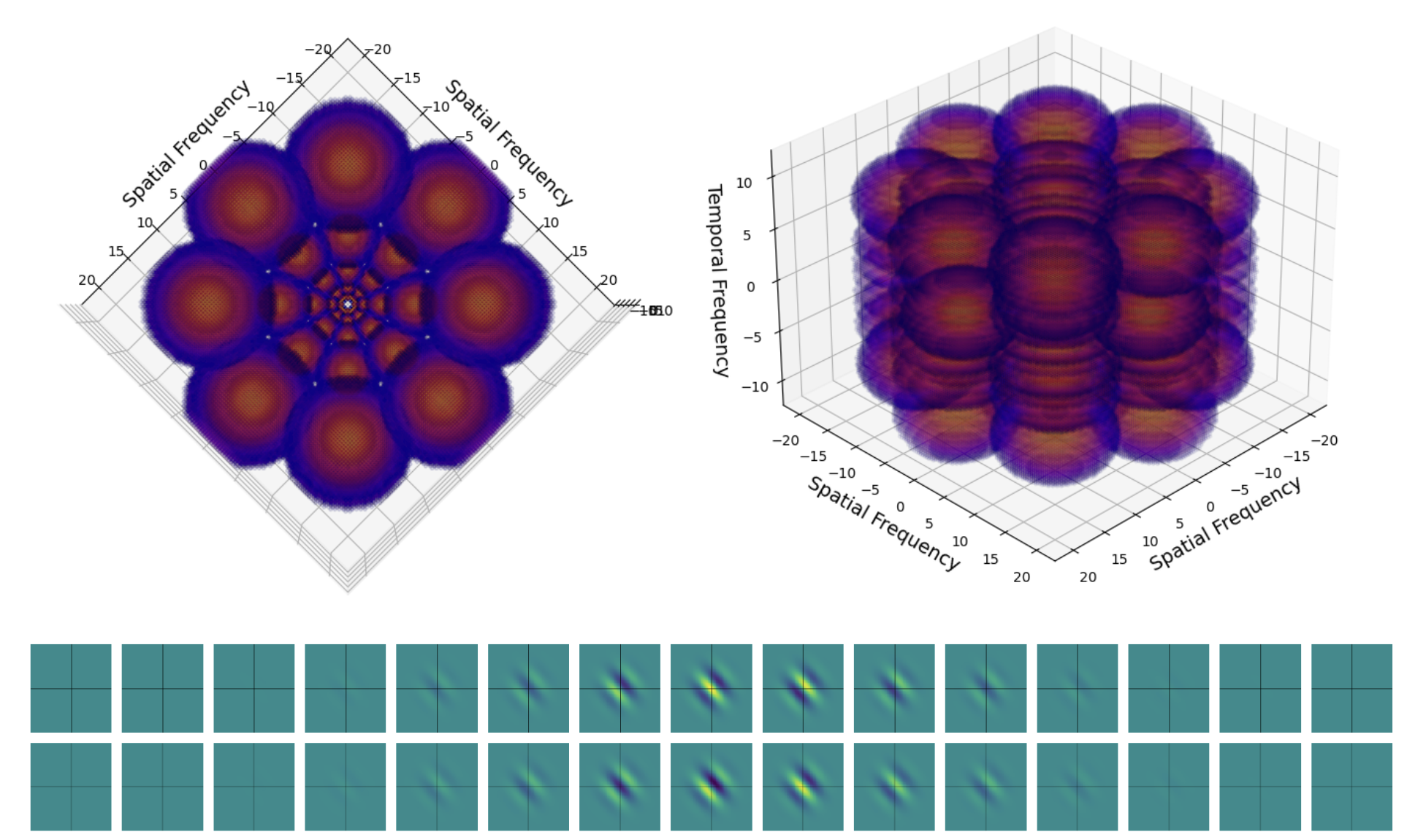}
  \caption{We design a filter bank of spatiotemporal Gabor filters that tile the spatiotemporal frequency space. Viewed  in Fourier space (top), filters are arranged in concentric rings that are stacked vertically, selecting for 8 orientations at increasing spatial and temporal frequencies. Each filter in the positive temporal frequency space has a quadrature pair mirrored across a line through the origin. Left: top view, Right: side view. Bottom: A real+imaginary pair of spatiotemporal filters, generated by inverse Fourier transform of the spatiotemporal frequency defined filters above. Left to right denotes the temporal axis, with frames sub-sampled for visualization. This pair of filters is sensitive to low spatial and low temporal frequency motion to the upper-right.}
  \label{fig:filterbank}
\end{figure}

The first step for expansion of our model into the temporal domain is the design of a human-vision informed spatiotemporal filter bank. We extended the classical multi-scale spatial pyramid \citep{heeger1995pyramid} into the time domain with a set of spatiotemporal filters centered around the spatiotemporal frequency space of human visual sensitivity. A subset of these filters in the Fourier domain with boundaries at half maximum amplitude is shown in Figure \ref{fig:filterbank} (top). When the inverse Fourier transform of a quadrature pair of frequency defined filters is taken, both a real and imaginary component is generated. These spatiotemporal filter pairs can be visualized in the pixel-frame space, enabling intuition for the type of motion a filter pair selects for. A sample of these filter pairs is shown in Figure \ref{fig:filterbank}, (bottom).

In the Fourier domain, our filter banks take the form of anisometric Gaussian functions centered on $(\mu_x, \mu_y, \mu_t)$ with variances $(\sigma_x^2, \sigma_y^2, \sigma_t^2)$

\begin{align}
    \tilde{f}(\mathbf{k}) \propto \exp\Bigg[-\frac{1}{2}\Bigg(
    &\left(\frac{k_x - \mu_x}{\sigma_x}\right)^2 
    + \left(\frac{k_y - \mu_y}{\sigma_y}\right)^2 \notag \\
    &+ \left(\frac{k_t - \mu_t}{\sigma_t}\right)^2
    \Bigg)\Bigg].
\end{align}

The filter positions are determined by the spatial scale ($\rho$), temporal scale ($\zeta$), and orientation $(\phi)$. These are defined in cylindrical coordinates as:

\begin{align}
    \mu_x &= \rho \cdot \cos(\phi)\\
    \mu_y &= \rho \cdot \sin(\phi)\\
    \mu_t &= \zeta
\end{align}

Finally, the standard deviations $\sigma_i$ scale with the centers via spatial and temporal scaling factors $\alpha_S = 0.4$ and $\alpha_T = 0.7$ such that:

\begin{align}
    \sigma_x &= \alpha_S \cdot \mu_x\\
    \sigma_y &= \alpha_S \cdot \mu_y\\
    \sigma_t &= \alpha_T \cdot \mu_t
\end{align}

For our filter bank, we sample the spatial scales, temporal scales, and orientations over the ranges:

\begin{align}
    \rho &\in \{1,2,4,8,16\}\\
    \zeta &\in \{1,2,4,8,16\}\\
    \phi &\in \{0,\frac{\pi}{4},\frac{\pi}{2},\frac{3\pi}{4},\pi,\frac{5\pi}{4},\frac{3\pi}{2},\frac{7\pi}{4}\}
\end{align}

We synthesize videos by matching on the Gramian of these filters. Unlike the spatial case, we find that for video textures, histogram matching over the whole image is needed in addition to statistics matching to achieve good syntheses.

We chose the range of the spatiotemporal space up to 16 CPD and 20 Hz, following the human spatiotemporal contrast sensitivity calculations for foveal vision \citep{kelly1979motion}, and spatiotemporal flicker fusion for peripheral vision \citep{krajancich2021perceptual}. Within this range, we designed filters as 3D Gaussians arranged in concentric stacked rings (shaped like a mochi-donut), similar to \citep{simoncelli1993distributed}. A single ring represents 8 filters of varying orientations (motion directions) at the same spatial and temporal frequency. Rings are arranged concentrically to select for increasing spatial frequencies, and stacked along the temporal axis to select for increasing temporal frequencies. Each filter located in the positive half of the temporal frequency space has a quadrature pair filter mirrored in the negative half of the temporal frequency space, in the opposite spatial quadrant. To tile the space, we chose a set of 8 orientations, 5 spatial scales, and 5 temporal scales, with a real and imaginary pair each, giving 200 oriented filters. We also include one spatial and one temporal low-pass filter, each with a real and imaginary pair, for 404 filter responses per colour channel and 1,212 in total. These spatiotemporal filter pairs correspond to a set of Gabor filters.

\FloatBarrier
\subsection{Spatiotemporal Statistics Families}

The benefit of interpretability of our method as pairwise correlation between filters is true for both the spatial-only filters as well as the spatiotemporal filters. For the spatiotemporal filters, given the much larger number of filters, we use a slightly different organizational approach, separating them by each parameter that varies. Table \ref{table:statsfamilies} organizes these statistics into families based on the relationship between the two filter images being correlated. We take specific note of the groups denoted by a single letter (C,O,S,T,P), as these correspond to correlations between filters that vary in only one attribute. These statistics are analogous to the hand-selected spatial statistics in previous pyramid-based texture and peripheral vision models including \citep{portilla2000parametric, freeman2011metamers, balas2006texture}. In addition, spatial texture investigations have shown that low-pass filter correlation statistics (LPS) are less important to successful synthesis, and correlation statistics across many filter attributes (such as COSTP) are most important \citep{dutell2023evaluating}.

\begin{table*}[!ht]
\centering
\renewcommand{\arraystretch}{1.15}
\small
\caption{Spatiotemporal correlation statistics grouped into families based on which filter attributes are cross-correlated (\emph{Cross}) versus held fixed (\emph{Same}). The five attributes are color channel, motion direction (orientation), spatial frequency, temporal frequency, and phase (real/imaginary). Special families (top): \textsc{auto} contains filter autocorrelations on the Gram diagonal; \textsc{lps}/\textsc{lpt} contain correlations involving spatial / temporal low-pass filters. Remaining families are partitioned by the number of attributes that vary, from one (e.g., C, O) to all five (\textsc{costp}).}
\begin{tabular}{@{}llllr@{}}
\toprule
\textbf{Group} & \textbf{Cross (varies)} & \textbf{Same (fixed)} & & \textbf{\#Stats} \\
\midrule
\multicolumn{5}{l}{\textit{Special families}} \\
\midrule
AUTO    & \multicolumn{2}{l}{Auto-correlations (Gram diagonal)}             & & 1{,}212   \\
LPS     & \multicolumn{2}{l}{Spatial low-pass correlations}                 & & 7{,}233   \\
LPT     & \multicolumn{2}{l}{Temporal low-pass correlations}                & & 7{,}233   \\
\midrule
\multicolumn{5}{l}{\textit{One attribute varies}} \\
\midrule
C       & color           & orientation, spatial, temporal, phase           & & 1{,}200   \\
O       & orientation     & color, spatial, temporal, phase                 & & 4{,}200   \\
S       & spatial         & color, orientation, temporal, phase             & & 2{,}400   \\
T       & temporal        & color, orientation, spatial, phase              & & 2{,}400   \\
P       & phase           & color, orientation, spatial, temporal           & & 600       \\
\midrule
\multicolumn{5}{l}{\textit{Two attributes vary}} \\
\midrule
CO      & color, orientation        & spatial, temporal, phase              & & 8{,}400   \\
CS      & color, spatial            & orientation, temporal, phase          & & 4{,}800   \\
CT      & color, temporal           & orientation, spatial, phase           & & 4{,}800   \\
CP      & color, phase              & orientation, spatial, temporal        & & 1{,}200   \\
OS      & orientation, spatial      & color, temporal, phase                & & 16{,}800  \\
OT      & orientation, temporal     & color, spatial, phase                 & & 16{,}800  \\
OP      & orientation, phase        & color, spatial, temporal              & & 4{,}200   \\
ST      & spatial, temporal         & color, orientation, phase             & & 9{,}600   \\
SP      & spatial, phase            & color, orientation, temporal          & & 2{,}400   \\
TP      & temporal, phase           & color, orientation, spatial           & & 2{,}400   \\
\midrule
\multicolumn{5}{l}{\textit{Three attributes vary}} \\
\midrule
COS     & color, orientation, spatial      & temporal, phase                & & 33{,}600  \\
COT     & color, orientation, temporal     & spatial, phase                 & & 33{,}600  \\
COP     & color, orientation, phase        & spatial, temporal              & & 8{,}400   \\
CST     & color, spatial, temporal         & orientation, phase             & & 19{,}200  \\
CSP     & color, spatial, phase            & orientation, temporal          & & 4{,}800   \\
CTP     & color, temporal, phase           & orientation, spatial           & & 4{,}800   \\
OST     & orientation, spatial, temporal   & color, phase                   & & 67{,}200  \\
OSP     & orientation, spatial, phase      & color, temporal                & & 16{,}800  \\
OTP     & orientation, temporal, phase     & color, spatial                 & & 16{,}800  \\
STP     & spatial, temporal, phase         & color, orientation             & & 9{,}600   \\
\midrule
\multicolumn{5}{l}{\textit{Four attributes vary}} \\
\midrule
COST    & color, orientation, spatial, temporal   & phase                   & & 134{,}400 \\
COSP    & color, orientation, spatial, phase      & temporal                & & 33{,}600  \\
COTP    & color, orientation, temporal, phase     & spatial                 & & 33{,}600  \\
CSTP    & color, spatial, temporal, phase         & orientation             & & 19{,}200  \\
OSTP    & orientation, spatial, temporal, phase   & color                   & & 67{,}200  \\
\midrule
\multicolumn{5}{l}{\textit{All five attributes vary}} \\
\midrule
COSTP   & color, orientation, spatial, temporal, phase   & ---              & & 134{,}400 \\
\midrule
\textbf{Total} & & & & \textbf{735{,}078} \\
\bottomrule
\end{tabular}
\label{table:statsfamilies}
\end{table*}

\FloatBarrier
\subsection{Video Syntheses}


Videos of all syntheses, including side-by-side comparisons with DyNCA, are available at \url{https://vdutell.github.io/gstn}.

We evaluate the quantitative performance on a variety of metrics, comparing GSTN video model to DyNCA, a recent model for video synthesis (Table \ref{tab:video_synthesis_quality_quant}). DyNCA leads on deep-feature appearance metrics and diversity match, while GSTN leads on temporal power-spectrum match and perceptual straightness, pixel-level metrics are near tied. We emphasize here that DyNCA is a fundamentally different type of model than GSTN with differing tradeoffs, optimized for synthesis quality and model size, and with a per-image/texture trained parameterization, that lacks interpretability and biological/perceptual connection.  

\begin{table}[h]
\centering
\scriptsize
\setlength{\tabcolsep}{3pt}
\caption{GSTN vs.\ DyNCA~\citep{pajouheshgar2023dynca} per video on five DynTex dynamic textures. Metrics are computed with identical code for both methods. Bold = better of the two (both when tied). LP-A/LP-V = LPIPS (AlexNet/VGG); Spec = L1 distance between normalized temporal power spectra; $|\Delta\mathrm{Div}|$ and $|\Delta\mathrm{Curv}|$ = absolute distance from the target's own frame diversity and perceptual-trajectory curvature~\citep{henaff2019perceptual}. SSIM, MSE and PSNR are alignment-based and weak for stochastic syntheses; per-video FID is noisy at small $n$ (read the mean). GSTN uses histogram projection every 50 evaluations, 3{,}000 evaluations.}
\label{tab:video_synthesis_quality_quant}
\begin{tabular}{@{}llrrrrrrrrrrr@{}}
\toprule
Video & Method & LP-A\,$\downarrow$ & LP-V\,$\downarrow$ & DISTS\,$\downarrow$ & SSIM\,$\uparrow$ & FID\,$\downarrow$ & CMMD\,$\downarrow$ & MSE\,$\downarrow$ & PSNR\,$\uparrow$ & Spec\,$\downarrow$ & $|\Delta\mathrm{Div}|$\,$\downarrow$ & $|\Delta\mathrm{Curv}|$\,$\downarrow$ \\
\midrule
grass & GSTN & 0.287 & 0.463 & 0.318 & 0.016 & 160 & 2.030 & \textbf{0.045} & \textbf{13.4} & \textbf{0.196} & 0.057 & \textbf{8.7} \\
 & DyNCA & \textbf{0.124} & \textbf{0.281} & \textbf{0.205} & \textbf{0.054} & \textbf{75} & \textbf{0.664} & 0.046 & \textbf{13.4} & 0.786 & \textbf{0.013} & 20.2 \\
\midrule
fire & GSTN & 0.557 & 0.602 & 0.347 & \textbf{0.398} & 120 & 2.156 & 0.014 & 18.7 & \textbf{0.152} & 0.117 & 8.2 \\
 & DyNCA & \textbf{0.140} & \textbf{0.334} & \textbf{0.182} & 0.356 & \textbf{31} & \textbf{1.696} & \textbf{0.013} & \textbf{18.8} & 0.902 & \textbf{0.090} & \textbf{3.9} \\
\midrule
fence & GSTN & 0.265 & 0.548 & 0.360 & \textbf{0.047} & 378 & 3.737 & 0.054 & 12.7 & \textbf{0.253} & 0.043 & \textbf{18.1} \\
 & DyNCA & \textbf{0.152} & \textbf{0.309} & \textbf{0.187} & 0.044 & \textbf{174} & \textbf{1.609} & \textbf{0.052} & \textbf{12.8} & 0.453 & \textbf{0.034} & 21.0 \\
\midrule
blooms & GSTN & 0.302 & 0.493 & 0.301 & \textbf{0.049} & 89 & 2.781 & 0.073 & 11.4 & \textbf{0.188} & 0.084 & 22.6 \\
 & DyNCA & \textbf{0.206} & \textbf{0.367} & \textbf{0.188} & 0.024 & \textbf{79} & \textbf{1.954} & \textbf{0.061} & \textbf{12.1} & 0.193 & \textbf{0.035} & \textbf{14.0} \\
\midrule
tree & GSTN & 0.202 & 0.467 & 0.295 & \textbf{0.061} & 64 & 1.501 & \textbf{0.032} & \textbf{15.0} & \textbf{0.094} & 0.025 & \textbf{4.7} \\
 & DyNCA & \textbf{0.174} & \textbf{0.321} & \textbf{0.234} & 0.057 & \textbf{54} & \textbf{0.770} & 0.046 & 13.4 & 0.254 & \textbf{0.012} & 29.8 \\
\midrule
\textbf{mean} & GSTN & 0.323 & 0.515 & 0.324 & \textbf{0.114} & 162 & 2.441 & \textbf{0.044} & \textbf{14.2} & \textbf{0.176} & 0.065 & \textbf{12.5} \\
 & DyNCA & \textbf{0.159} & \textbf{0.323} & \textbf{0.199} & 0.107 & \textbf{83} & \textbf{1.339} & \textbf{0.044} & 14.1 & 0.518 & \textbf{0.037} & 17.8 \\
\bottomrule
\end{tabular}
\end{table}

\FloatBarrier
\subsection{LLM Use}
The GSTN model itself was created without the use of LLMs. Generation of the code for foveated syntheses was done without the use of LLMs. Code to run, manage, and queue parameter sweeps and depletion runs on the GPU was generated using LLMs. Code to aggregate synthesis images, calculate quantitative metrics, and plot summary figures was generated using LLMs.

\FloatBarrier
\section{Full Qualitative Comparison on DTD-141}
\label{app:dtd141}

For completeness we include every image in the evaluation set below: all 141 images (47 categories, $n=3$ each). This is the complete subset of the DTD dataset used in quantitative comparisons throughout the manuscript except where otherwise specified.

\begin{figure}[h]
\caption{\textbf{Full qualitative comparison over the DTD-141 evaluation set} (next 9 pages).}
\label{fig:dtd141}
\end{figure}

\includepdf[pages=-,pagecommand={\thispagestyle{plain}}]{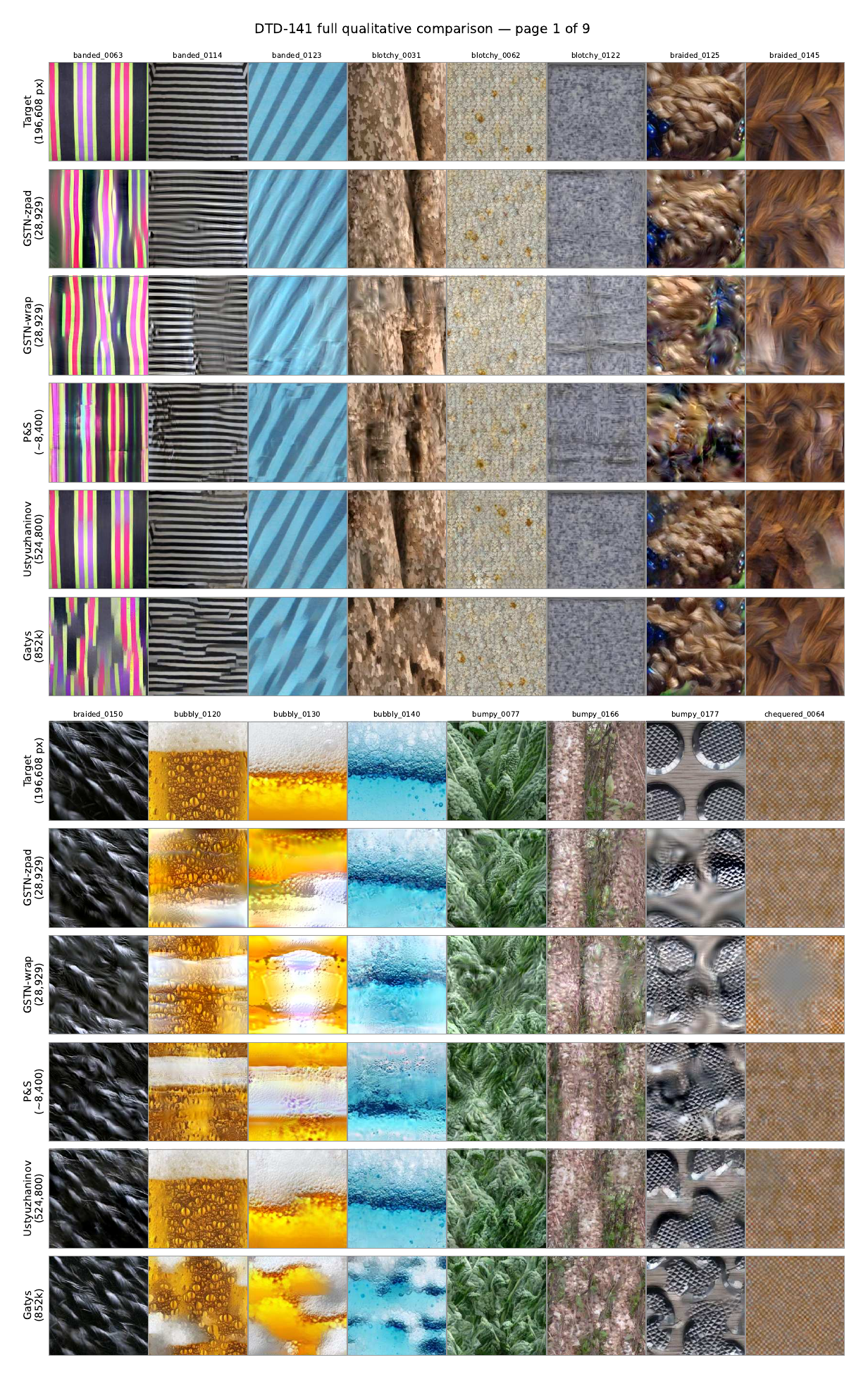}

\newpage
\clearpage

\end{document}